\documentclass[twoside]{article}

\PassOptionsToPackage{numbers, compress}{natbib}

\usepackage[accepted]{aistats2021}
\usepackage{amsfonts}
\usepackage[T1]{fontenc}
\usepackage[utf8]{inputenc}
\usepackage{tabularx,ragged2e,booktabs,caption}
\newcolumntype{C}[1]{>{\Centering}m{#1}}

\usepackage[utf8]{inputenc}
\usepackage[T1]{fontenc}
\usepackage{url}
\usepackage{booktabs}
\usepackage{amsfonts, amsthm, amsmath}
\usepackage{mathtools}
\usepackage{nicefrac}
\usepackage{microtype}

\usepackage{natbib}

\usepackage{graphicx}
\usepackage[labelfont=bf]{caption}
\usepackage[format=hang]{subcaption}
\usepackage{wrapfig}

\usepackage[usenames,dvipsnames]{xcolor}
\definecolor{shadecolor}{gray}{0.9}

\usepackage{ragged2e}

\usepackage{algorithm, algorithmic}

\usepackage[colorlinks,linktoc=all]{hyperref}
\usepackage[all]{hypcap}
\hypersetup{citecolor=Violet}
\hypersetup{linkcolor=black}
\hypersetup{urlcolor=MidnightBlue}

\usepackage[nameinlink]{cleveref}

\usepackage[acronym,smallcaps,nowarn]{glossaries}
\glsdisablehyper{}

\newacronym{RL}{rl}{reinforcement learning}
\newacronym{EM}{em}{expectation maximization}
\newacronym{IS}{is}{importance sampling}

\newacronym{HER}{her}{hindsight experience replay}
\newacronym{HPG}{hpg}{hindsight policy gradient}
\newacronym{hEM}{$\mathrm{h}$em}{hindsight expectation maximization}

\newacronym{VAE}{vae}{variational auto-encoders}

\newacronym{TD}{td}{temporal difference}

\newacronym{hPI}{hpi}{hindsight policy improvement}

\newacronym{MDP}{mdp}{{M}arkov decision process}
\newacronym{Variational RL}{variational rl}{variational reinforcement learning}

\newacronym{ELBO}{elbo}{evidence lower bound}

\usepackage{tikz}
\usetikzlibrary{bayesnet}

\pgfdeclarelayer{edgelayer}
\pgfdeclarelayer{nodelayer}
\pgfsetlayers{edgelayer,nodelayer,main}

\definecolor{hexcolor0xbfbfbf}{rgb}{0.749,0.749,0.749}

\tikzset{>=latex}
\tikzstyle{none}   = [inner sep=0pt]
\tikzstyle{line}   = [ -, thick, shorten <=1pt, shorten >=1pt ]
\tikzstyle{arrow}  = [ ->, thick, shorten <=1pt, shorten >=1pt ]
\tikzstyle{ardash} = [ dashed, ->, thick, shorten <=1pt, shorten >=1pt ]

\tikzstyle{empty}=[circle,opacity=0.0,text opacity=1.0,inner sep=0pt]
\tikzstyle{box}=[rectangle,fill=White,draw=Black]
\tikzstyle{filled}=[circle,thick,fill=hexcolor0xbfbfbf,draw=Black]
\tikzstyle{hollow}=[circle,thick,fill=White,draw=Black]
\tikzstyle{param}=[rectangle,fill=Black,draw=Black,inner sep=0pt,minimum width=4pt,minimum height=4pt]
\tikzstyle{paramhollow}=[rectangle,thick,fill=White,draw=Black,inner sep=0pt,minimum
width=4pt,minimum height=4pt]

\newtheorem{theorem}{Theorem}
\newtheorem{proposition}{Proposition}

\begin{document}

%

%

\twocolumn[

\aistatstitle{Hindsight Expectation Maximization for Goal-conditioned Reinforcement Learning}

\aistatsauthor{ Yunhao Tang \And Alp Kucukelbir }

\aistatsaddress{ Columbia University \And  Columbia University \& Fero Labs } ]

\begin{abstract}
We propose a graphical model framework for goal-conditioned \gls{RL}, with an \gls{EM} algorithm that operates on the
lower bound of the \gls{RL} objective. The E-step provides a natural interpretation of how `learning in hindsight'
techniques, such as \gls{HER}, handle extremely sparse goal-conditioned rewards. The M-step reduces policy
optimization to supervised learning updates, which stabilizes end-to-end training on high-dimensional inputs
such as images. Our proposed method, called \gls{hEM}, significantly outperforms model-free baselines on a wide
range of goal-conditioned benchmarks with sparse rewards.
\end{abstract}

\glsresetall
\section{Introduction}

In goal-conditioned \gls{RL}, an agent seeks to achieve a goal through interactions with the environment. At each step,
the agent receives a reward, which ideally reflects how well it is achieving its goal. Traditional
\gls{RL} methods leverage these rewards to learn good policies. As such, the effectiveness of these methods rely on how
informative the rewards are.

This sensitivity of traditional \gls{RL} algorithms has led to a flurry of activity around reward shaping \citep{ng1999policy}.
This limits the applicability of \gls{RL}, as reward shaping is often specific to an environment and task --- a practical obstacle to wider applicability. Binary rewards,
however, are trivial to specify. The agent receives a strict indicator of success when it has achieved its goal; until
then, it receives precisely zero reward. Such a sparsity of reward signals renders goal-conditioned \gls{RL} extremely
challenging for traditional methods \citep{andrychowicz2017hindsight}.

How can we navigate such binary reward settings? Consider an agent that explores its environment but fails to achieve
its goal. One idea is to treat, \emph{in hindsight}, its exploration as having achieved some other goal. By relabeling
a `failure' relative to an original goal as a `success' with respect to some other goal, we can imagine an agent succeeding
frequently at many goals, in spite of failing at its original goals. This insight motivates \gls{HER}
\citep{andrychowicz2017hindsight}, an intuitive strategy that enables off-policy \gls{RL} algorithms, such as
\citep{mnih2013,lillicrap2015continuous}, to function in sparse binary reward settings.

The statistical simulation of rare events occupies a similar setting. Consider estimating an expectation of low-probability events using Monte Carlo sampling. The variance of this estimator relative to its expectation is too high to be practical \citep{rubino2009rare}. A powerful approach to reduce variance is \gls{IS} \citep{casella2002statistical}. The idea is to adapt the sampling procedure such that these rare events occur frequently, and then to adjust the final computation. Could \gls{IS} help in binary reward \gls{RL} settings too?

\begin{figure*}[!t]
\centering
\subcaptionbox{\textbf{Point mass}}[.24\linewidth]{\includegraphics[width=1.5in]{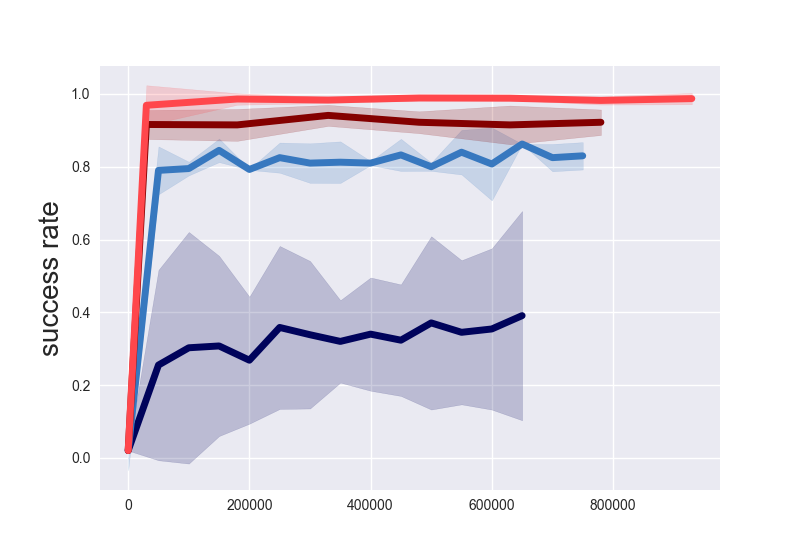}}
\subcaptionbox{\textbf{Reacher goal}}[.24\linewidth]{\includegraphics[width=1.5in]{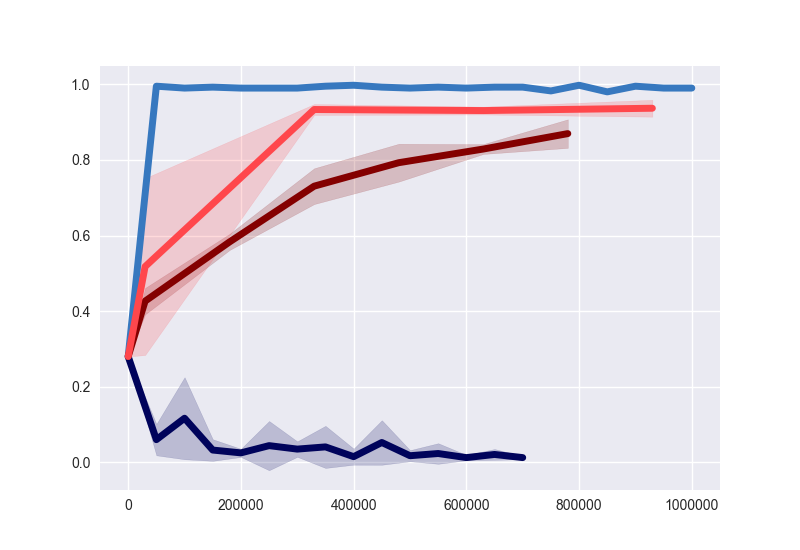}}
\subcaptionbox{\textbf{Fetch robot}}[.24\linewidth]{\includegraphics[width=1.5in]{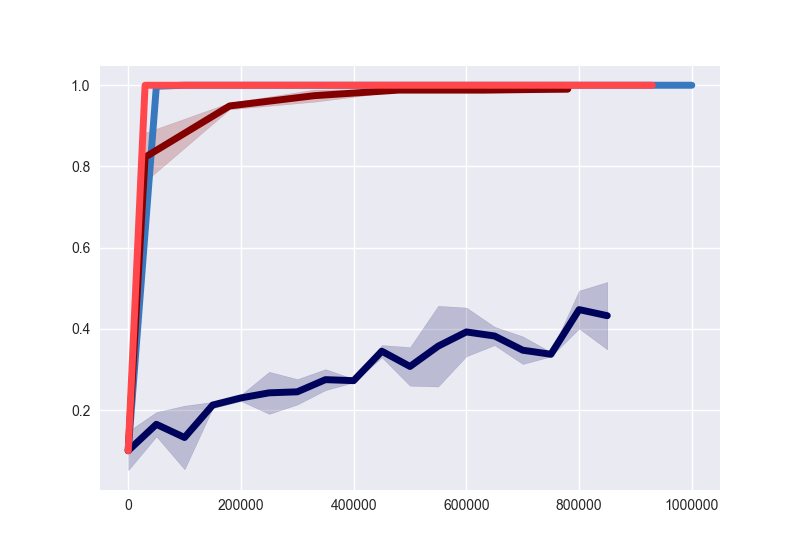}}
\subcaptionbox{\textbf{Sawyer robot}}[.24\linewidth]{\includegraphics[width=1.5in]{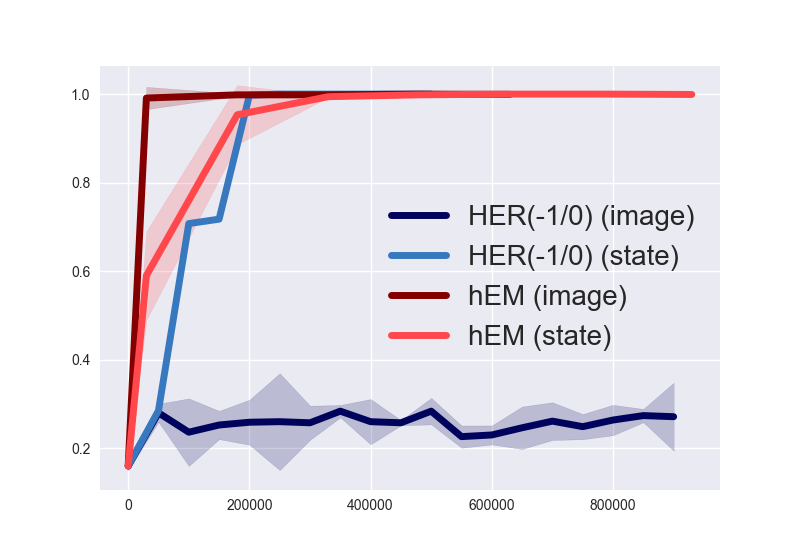}}
\caption{\small{Training curves of \gls{hEM} and \gls{HER} on four goal-conditioned \gls{RL} benchmark tasks. Inputs are either state-based or image-based. The y-axis shows the success rates and the x-axis shows the training time steps. All curves are calculated based on averages over $5$ random seeds. Here, \gls{HER}($-1/0$) denotes \gls{HER} trained on rewards $r=-\mathbb{I}[\text{failure}]$, see \Cref{sec:experiments} for details.  \gls{hEM} consistently performs better than or similar to \gls{HER} across all tasks. The performance gains are significant for high-dimensional image-based tasks.}}
\label{figure:intro}
\end{figure*}

\glsreset{hEM}
\paragraph{Main idea.} We propose a probabilistic framework for goal-conditioned \gls{RL} that formalizes the intuition of hindsight using ideas from statistical simulation. We equate the traditional \gls{RL} objective to maximizing the evidence of our probabilistic model. This leads to a new algorithm, \gls{hEM}, which maximizes a tractable lower bound of the original objective \citep{blei2017}. A central insight is that the E-step naturally interprets hindsight replay as a special case of \gls{IS}.

\Cref{figure:intro} compares \gls{hEM} to \gls{HER} \citep{andrychowicz2017hindsight} on four goal-conditioned \gls{RL} tasks with low-dimensional state and high-dimensional images as inputs. While \gls{hEM} performs consistently well on both inputs, \gls{HER} struggles with image-based inputs. This is due to how \gls{HER} leverages hindsight replay within a \gls{TD}-learning procedure; performance degrades sharply with the dimensionality of the inputs (as observed previously in \citep{lee2019stochastic,kostrikov2020image}; also see \Cref{sec:experiments}). In contrast, \gls{hEM} leverages hindsight experiences through the lens of \gls{IS}, thus enabling better performance in high dimensions.




The rest of this section presents a quick background on goal-conditioned \gls{RL} and probabilistic inference. Expert readers may jump ahead to \Cref{label:hEM}.

\paragraph{Goal-conditioned \gls{RL}.}
Consider an agent that interacts with an environment in episodes.
At the beginning of each episode, a goal $g \in \mathcal{G}$ is fixed.
At a discrete time $t \geq 0$, an agent in state $s_t \in \mathcal{S}$ takes action $a_t \in \mathcal{A}$, receives a reward $r(s_t,a_t,g) \in \mathbb{R}$, and transitions to its next state $s_{t+1} \sim p(\cdot \mid s_t,a_t)\in \mathcal{S}$.
This process is independent of goals.
A policy $\pi(a \mid s,g):\mathcal{S}\times\mathcal{G}\mapsto \mathcal{P}(\mathcal{A})$ defines a map from state and goal to distributions over actions. Given a distribution over goals $g \sim p(\cdot)$, we consider the undiscounted episodic return $J(\pi) \coloneqq \mathbb{E}_{g\sim p(\cdot)}\big[\mathbb{E}_\pi[\sum_{t=0}^{T-1}  r(s_t,a_t,g)]\big]$.


\paragraph{Probabilistic inference.}
Consider data as observed random variables $x \in \mathcal{X}$. Each measurement $x$ is a discrete or continuous random
variable. A likelihood $p_\theta(x \mid z)$ relates each measurement to latent variables $z\in\mathcal{Z}$ and unknown,
but fixed, parameters $\theta$. The full probabilistic generative model specifies a prior over the latent variable
$p(z)$. Bayesian inference requires computing the posterior $p(z \mid x)$ --- an intractable task for all but a small
class of simple models.

Variational inference approximates the posterior by matching a tractable density $q_\phi(z \mid x)$ to the posterior. The following calculation specifies this procedure:
\begin{align}
\log p(x)
&= \log \mathbb{E}_{z\sim p(\cdot)} \left[ p_\theta(x \mid z) \right] \nonumber \\
&=
 \log \mathbb{E}_{z\sim q_\phi(\cdot \mid x)}
 \left[
 p_\theta(x \mid z)\frac{p(z)}{q_\phi(z \mid x)}
 \right] \nonumber \\
 &\geq
 \mathbb{E}_{z\sim q_\phi(\cdot \mid x)}
 \left[
 \log p_\theta(x \mid z)\frac{p(z)}{q_\phi(z \mid x)}
 \right]  \label{eq:vae-is} \\
&=
 \mathbb{E}_{z\sim q_\phi(\cdot \mid x)}
 \left[
 \log p_\theta(x \mid z)
 \right] - \mathbb{KL}[q_\phi(\cdot\mid z) \;\|\; p(z)]
 \nonumber \\
 &\eqqcolon
 L(p_\theta,q). \label{eq:vae-elbo}
\end{align}

\begin{figure*}[!t]
\centering
\subcaptionbox{\small{Probabilistic inference}}[.25\linewidth]{
\begin{tikzpicture}
	\begin{pgfonlayer}{nodelayer}
		\node [style=filled] (0) at (0, 0) {$x$};
		\node [style=hollow] (1) at (0, 1.5) {$z$};
		\node [style=box] (4) at (-1.2, 1.5) {$\theta$};
		\node [style=box] (5) at (1.2, 1.5) {$\phi$};
		\node [style=empty] (10) at (0, 0.66) {};
		\node [style=empty] (11) at (-0.5, 0.66) {};
		\node [style=empty] (12) at (0.5, 0.66) {};
	\end{pgfonlayer}
	\begin{pgfonlayer}{edgelayer}
		\draw [style=arrow] (1) to (0);
		\draw [style=arrow] (4) to (1);
		\draw [style=arrow][bend right=60,dashed] (0) to (1);
		\draw [style=arrow][dashed] (5) to (1);
	\end{pgfonlayer};
\end{tikzpicture}
}
\subcaptionbox{\small{Variational \textsc{rl}}}[.23\linewidth]{
\begin{tikzpicture}
	\begin{pgfonlayer}{nodelayer}
		\node [style=filled] (0) at (0, 0) {$O$};
		\node [style=hollow] (1) at (0, 1.5) {$\tau$};
		\node [style=box] (4) at (-1.2, 1.5) {$\theta$};
		\node [style=box] (5) at (1.2, 1.5) {$q$};
		\node [style=empty] (10) at (0, 0.66) {};
		\node [style=empty] (11) at (-0.5, 0.66) {};
		\node [style=empty] (12) at (0.5, 0.66) {};
	\end{pgfonlayer}
	\begin{pgfonlayer}{edgelayer}
		\draw [style=arrow] (1) to (0);
		\draw [style=arrow] (4) to (1);
		\draw [style=arrow][dashed] (5) to (1);
	\end{pgfonlayer};
\end{tikzpicture}
}
\subcaptionbox{\small{Goal-conditioned \textsc{rl} \\ (Generative model)}}[.23\linewidth]{
\begin{tikzpicture}
	\begin{pgfonlayer}{nodelayer}
		\node [style=filled] (0) at (0, 0) {$O$};
		\node [style=hollow] (1) at (0, 1.5) {$\tau$};
		\node [style=hollow] (4) at (-1.2, 1.5) {$g$};
		\node [style=box] (5) at (-1.2, 0) {$\theta$};
		\node [style=empty] (10) at (0, 0.66) {};
		\node [style=empty] (11) at (-0.5, 0.66) {};
		\node [style=empty] (12) at (0.5, 0.66) {};
	\end{pgfonlayer}
	\begin{pgfonlayer}{edgelayer}
		\draw [style=arrow] (1) to (0);
		\draw [style=arrow] (4) to (1);
		\draw [style=arrow] (5) to (1);
		\draw [style=arrow] (4) to (0);
	\end{pgfonlayer};
\end{tikzpicture}
}
\subcaptionbox{\small{Goal-conditioned \textsc{rl} \\ (Inference model)}}[.23\linewidth]{
\begin{tikzpicture}
	\begin{pgfonlayer}{nodelayer}
		\node [style=hollow] (1) at (0, 1.5) {$\tau$};
		\node [style=hollow] (4) at (-1.2, 1.5) {$g$};
		\node [style=box] (5) at (-0.6, 0) {$q$};
		\node [style=empty] (10) at (0, 0.66) {};
		\node [style=empty] (11) at (-0.5, 0.66) {};
		\node [style=empty] (12) at (0.5, 0.66) {};
	\end{pgfonlayer}
	\begin{pgfonlayer}{edgelayer}
		\draw [style=arrow][dashed] (5) to (1);
		\draw [style=arrow][dashed] (5) to (4);
	\end{pgfonlayer};
\end{tikzpicture}
}
\caption{\small{Graphical models. Solid lines represent generative models and dashed lines represent inference models. Circles represent random variables and squares represent parameters. Shading indicates that the random variable is observed.}}
\label{figure:rlasinference}
\end{figure*}
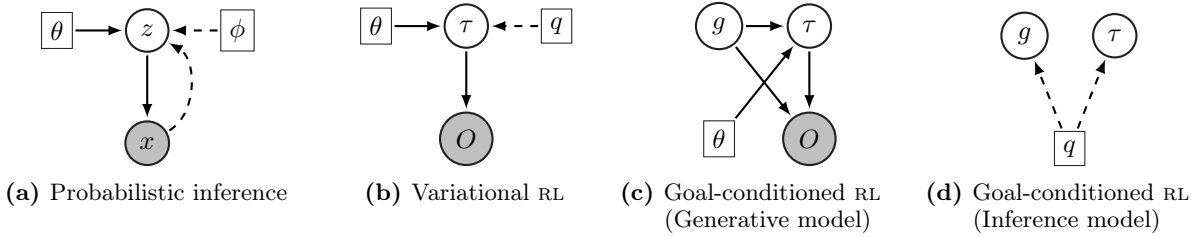

(For a detailed derivation, please see \citep{blei2017}.) \Cref{eq:vae-elbo} defines the \gls{ELBO} $L(p_\theta,q)$. Matching the tractable density $q_\phi(z \mid x)$ to the posterior thus turns into maximizing the \gls{ELBO} via \gls{EM} \citep{moon1996expectation} or stochastic gradient ascent \citep{kingma2013auto, ranganath2014black}.
\Cref{figure:rlasinference}(a) presents a graphical model of the above.
For a fixed set of $\theta$ parameters, the optimal variational distribution $q$ is the true posterior $\arg\max_q L(p_\theta,q) \equiv p(z\mid x) \coloneqq p(z) p_\theta(x \mid z) / p(x)$. From an \gls{IS} perspective, note how the variational distribution $q_\phi(z \mid x)$ serves as a proposal distribution in place of $p(z)$ in the derivation of \Cref{eq:vae-is}. 

\section{A Probabilistic Model for Goal-conditioned Reinforcement Learning}
\label{label:hEM}

Probabilistic modeling and control enjoy strong connections, especially in linear systems
\citep{kalman1959general,todorov2008general}. Two recent frameworks connect probabilistic inference to general \gls{RL}: Variational \gls{RL} \citep{kober2009policy,levine2013variational,abdolmaleki2018maximum,song2019v} and \gls{RL} as inference \citep{ziebart2010modeling,haarnoja2018latent,levine2018reinforcement}. We situate our probabilistic model by first presenting Variational \gls{RL} below. (\Cref{appendix:graphicalmodels} presents a detailed comparison to \gls{RL} as inference.)

\paragraph{Variational \gls{RL}.} Begin by defining a trajectory random variable $\tau \equiv (s_t,a_t)_{t=1}^{T-1}$ to encapsulate a sequence of state and action pairs. The random variable is generated by a factorized distribution $a_t\sim \pi_\theta(\cdot \mid s_t),s_{t+1}\sim p(\cdot \mid s_t,a_t)$, which defines the joint distribution $p_\theta(\tau) \coloneqq \Pi_{t=0}^{T-1} \pi_\theta(a_t \mid s_t) \; p(s_{t+1} \mid s_t,a_t)$.
Conditional on $\tau$, define the distribution of a binary optimality variable as $p(O=1 \mid \tau) \propto \exp(\sum_{t=0}^{T-1} r(s_t,a_t) / \alpha)$ for some $\alpha > 0$, where we assume $r(s_t,a_t) \geq 0$ without loss of generality. Optimizing the standard \gls{RL} objective corresponds to maximizing the evidence of $\log p(O=1)$, where all binary variables are treated as observed and equal to one. Positing a variational approximation to the posterior over trajectories gives the following lower bound,
\begin{align}
    \log p(O=1)
    \geq
    &\; \mathbb{E}_{q(\tau)}
    \left[
        \log p(O=1 \mid \tau)
    \right]
    - \nonumber \\ &\mathbb{KL} \left[q(\tau) \;\|\; p_\theta(\tau) \right] \eqqcolon L(\pi_\theta, q). \label{eq:rl-elbo}
\end{align}
\Cref{figure:rlasinference}(b) shows a combined graphical model for both the generative and inference models of Variational \gls{RL}. \Cref{eq:rl-elbo} is typically maximized using \gls{EM} (e.g., \citep{peters2010relative,abdolmaleki2018maximum,song2019v}), by alternating updates between $\theta$ and $q(\tau)$. Note that Variational \gls{RL} does not model goals.

\subsection{Probabilistic goal-conditioned reinforcement learning}

To extend the Variational \gls{RL} framework to incorporate goals, introduce a goal variable $g$ and a prior distribution
$g\sim p(\cdot)$. Conditional on a goal $g$, the trajectory variable $\tau
\equiv (s_t,a_t)_{t=0}^{T-1} \sim p(\cdot \mid \theta, g)$ is sampled by executing the policy $\pi_\theta(a \mid s, g)$ in the
\gls{MDP}. Similar to Variational \gls{RL}, the joint distribution factorizes as $p(\tau \mid \theta,g) \coloneqq
\Pi_{t=0}^{T-1} \pi_\theta(a_t \mid s_t,g) p(s_{t+1} \mid s_t,a_t)$. Now, define a goal-conditioned binary optimality
variable $O$, such that $p(O=1 \mid \tau,g) \coloneqq R(\tau,g) \coloneqq \sum_{t=0}^{T-1} r(s_t,a_t,g) / \alpha$ where
$\alpha > 0$ normalizes this density. \Cref{figure:rlasinference}(c) shows a graphical model of just this generative
model. Treat the optimality variables as the observations and assume $O \equiv 1$. The following proposition shows the
equivalence between inference in this model and traditional goal-conditioned \gls{RL}.
\begin{proposition}
\label{prop:equiv}
(Proof in \Cref{appendix:proof-quiv}.) Maximizing the evidence of the probabilistic model is equivalent to maximizing returns in the goal-conditioned \gls{RL} problem, i.e.,
\begin{align}
    \arg\max_\theta \log p(O=1) = \arg\max_\theta J(\pi_\theta).
    \label{eq:equiv}
\end{align}
\end{proposition}

\subsection{Challenges with direct optimization}
\Cref{eq:equiv} implies that algorithms that maximize the evidence of such probabilistic models could be readily applied to goal-conditioned \gls{RL}. Unlike typical probabilistic inference settings, the evidence here can technically be directly optimized.
Indeed, $p(O=1) \equiv J(\pi_\theta)$ could be maximized via traditional \gls{RL} approaches e.g., policy gradients \citep{rauber2017hindsight}. In particular, the \textsc{reinforce} gradient estimator \citep{williams1992} of \Cref{eq:equiv} is given by as $\eta_\theta = \sum_{t\geq 0}  \sum_{t^\prime \geq t}  r(s_{t^\prime},a_{t^\prime},g) \nabla_\theta \log \pi_\theta(a_{t} \mid s_{t},g) \approx \nabla_\theta J(\pi_\theta)$, where $g\sim p(\cdot)$ and $(s_t,a_t)_{t=0}^{T-1}$ are sampled on-policy. The direct optimization of $\log p(O=1)$ consists of a gradient ascent sequence $\theta \leftarrow \theta + \eta_\theta$. However, this poses a practical challenge in goal-conditioned \gls{RL}. To see why, consider the following example.

\paragraph{Illustrative example.}  Consider a one-step \gls{MDP} with $T=1$ where $\mathcal{S} = \{s_0\}$, $\mathcal{A} = \mathcal{G}$, $r(s,a,g) = \mathbb{I}[a=g]$. Assume that there are a finite number of actions and goals $|\mathcal{A}| = |\mathcal{G}| = k$.

The following theorem shows the difficulty in building a practical estimator for $\nabla_\theta J(\pi_\theta)$.

\begin{theorem}
\label{thm:sparsity}
(Proof in \Cref{appendix:proof-sparsity}.) Consider the example above.  Let the policy $\pi_\theta(a\mid s,g) = \text{softmax}(L_{a,g})$ be parameterized by logits $L_{a,g}$ and let $\eta_{a,g}$ be the one-sample \textsc{reinforce} gradient estimator of $L_{a,g}$. Assume a uniform distribution over goals $p(g) = 1/k$ for all $g\in \mathcal{G}$. Assume that the policy is randomly initialized (e.g. $L_{a,g}\equiv L,\forall a,g$ for some $L$). Let $\text{MSE}[x]$ be the mean squared error $\text{MSE}[x] \coloneqq \mathbb{E}[(x - \mathbb{E}[\eta_{a,g}])^2]$. It can be shown that the relative error of the estimate   $\sqrt{\text{MSE}[\eta_{a,g}]} / \mathbb{E}[\eta_{a,g}] = k (1 + o(1))$ grows approximately linearly with $k$ for all $\forall a\in \mathcal{A},g\in \mathcal{G}$.
\end{theorem}

The above theorem shows that in the simple setup above, the relative error of the \textsc{reinforce} gradient estimator grows linearly in $k$. This implies that to reduce the error with traditional Monte Carlo sampling would require $o(k^2)$ samples, which quickly becomes intractable as $k$ increases. Though variance reduction methods such as control variates \citep{sutton1999} could be of help, it does not change the sup-linear growth rate of samples (see comments in \Cref{appendix:proof-sparsity}). The fundamental bottleneck is that dense  gradients, where $r(s,a,g) \nabla_\theta \log \pi_\theta(a\mid s,g) \neq 0$, are rare events with probability $\nicefrac{1}{k}$, which makes them difficult to estimate with on-policy measures \citep{rubino2009rare}. This example hints at similar issues with more realistic cases and motivates an \gls{IS} approach to address the problem.


\subsection{Tractable lower bound}

Consider a variational inequality similar to \Cref{eq:vae-elbo} with a variational distribution $q(\tau,g)$
\begin{align}
\begin{split}
    \log p(O=1)
    &= \log \mathbb{E}_{q(\tau,g)}
    \left[
    p(O=1\mid\tau,g) \frac{p(g)p(\tau\mid\theta,g)}{q(\tau,g)}
    \right] \\
    &\geq \mathbb{E}_{q(\tau,g)}
    \left[
    \log p(O=1\mid\tau,g) \frac{p(g)p(\tau\mid\theta,g)}{q(\tau,g)}
    \right] \\
    &= \mathbb{E}_{q(\tau,g)}
    \left[\log p(O=1\mid\tau,g)\right] - \\
    &\ \ \ \ \mathbb{KL}\left[q(\tau,g) \;\|\; p(g) p(\tau \mid \theta,g)\right] \\
    &\eqqcolon L(\pi_\theta,q). \label{eq:vae-rl-elbo}
\end{split}
\end{align}
This variational distribution corresponds to the inference model in \Cref{figure:rlasinference}(d). As with typical graphical models, instead of maximizing $\log p(O=1)$, consider maximizing its \gls{ELBO} $L(\pi_\theta,q)$ with respect to both $\theta$ and variational distribution $q(\tau,g)$. Our key insight lies in the following observation: the bottleneck of the direct optimization of $\log p(O=1)$ lies in the sparsity of $p(O=1\mid\tau,g) = \sum_{t=0}^{T-1} r(s_t,a_t,g)$, where $(\tau,g)$ are sampled with the on-policy measure $g\sim p(\cdot),\tau\sim p(\cdot\mid\theta,g)$. The variational distribution $q(\tau,g)$ serves as a \gls{IS} proposal in place of $p(\tau\mid\theta,g)p(g)$. If $q(\tau,g)$ puts more probability mass on $(\tau,g)$ pairs with high returns (high $p(O=1\mid\tau,g)$), the rewards become dense and learning becomes feasible. In the next section, we show how hindsight replay \citep{andrychowicz2017hindsight} provides an intuitive and effective way to select such a $q(\tau,g)$.

\section{Hindsight Expectation Maximization}
\label{label:algo}

The \gls{EM}-algorithm \citep{moon1996expectation}
 for \Cref{eq:vae-rl-elbo} alternates between an E- and M-step: at iteration $t$, denote the policy parameter to be $\theta_t$ and the variational distribution to be $q_t$.
\begin{align}
    &\text{E-step:}\  q_{t+1} = \arg\max_q L(\pi_{\theta_t},q),\ \nonumber \\ &\text{M-step:} \ \theta_{t+1} = \arg\max_\theta L(\pi_\theta,q_{t+1}). \label{eq:alternate}
\end{align}
This ensures a monotonic improvement in the \gls{ELBO} $L(\pi_{\theta_{t+1}},q_{t+1}) \geq L(\pi_{\theta_t},q_t)$. We discuss these two alternating steps in details below, starting with the M-step.

\paragraph{M-step: Optimization for $\pi_\theta$.}
Fixing the variational distribution $q(\tau,g)$, to optimize $L(\pi_\theta,q)$ with respect to $\theta$ is equivalent to
\begin{align}
    &\max_\theta \mathbb{E}_{q(\tau,g)}
    \left[
    \log p(\tau\mid\theta,g)
    \right] \nonumber \\
    &\equiv
    \max_\theta \mathbb{E}_{q(\tau,g)}
    \left[\sum_{t=0}^{T-1} \log \pi_\theta(a_t \mid s_t,g)\right]. \label{eq:pi-update}
\end{align}
The right hand side of \Cref{eq:pi-update} corresponds to a supervised learning problem where learning samples come from $q(\tau,g)$. Prior studies have adopted this idea and developed policy optimization algorithms in this direction \citep{abdolmaleki2018maximum,song2019v,silver2017mastering,vuong2018supervised,schrittwieser2019mastering}. In practice, the M-step is carried out partially where $\theta$ is updated with gradient steps instead of optimizing \Cref{eq:pi-update} fully.

\paragraph{E-step: Optimization for $q(\tau,g)$.}
The choice of  $q(\tau,g)$ should satisfy two desirable properties: \textbf{(P.1)} it leads to monotonic improvements in $\log p(O=1) \equiv J(\pi_\theta)$ or a lower bound thereof; \textbf{(P.2)} it provide dense learning signals for the M-step. The posterior distribution $p(\tau,g\mid O=1)$ achieves \textbf{(P.1)} and \textbf{(P.2)} in a near-optimal way, in that it is the maximizer of the E-step in \Cref{eq:alternate}, which monotonically improves the \gls{ELBO}. The posterior also provides dense reward signals to the M-step because  $p(\tau,g\mid O=1)\propto p(O=1\mid\tau,g)$. In practice, one chooses a variational distribution $q(\tau,g)$ as an alternative to the intractable posterior by maximizing \Cref{eq:vae-rl-elbo}. Below, we show it is possible to achieve (\textbf{P.1})(\textbf{P.2}) even though the E-step is not carried out fully. By plugging in $p(O=1\mid\tau,g) = \sum_{t=0}^{T-1} r(s_t,a_t,g) / \alpha$, we write the \gls{ELBO} as
\begin{align}
    L(\pi_\theta,q)
    =
    &\underbrace{
    \mathbb{E}_{q(\tau,g)}
        \Bigg[
        \frac{\sum_{t=0}^{T-1} r(s_t,a_t,g)}{\alpha}
        \Bigg]
    }_{\text{first\ term}}
    \,
    +
    \, \nonumber \\
    &\underbrace{
    \vphantom{\Bigg[}
    -\mathbb{KL}[q(\tau,g) \;\|\; p(g) p(\tau\mid\theta,g)]
    }_{\text{second\ term}}. \label{eq:e-step}
\end{align}
We now examine alternative ways to select the variational distribution $q(\tau,g)$.

\paragraph{Prior work.} State-of-the-art model-free algorithms such as \textsc{mpo} \citep{abdolmaleki2018maximum,song2019v} applies a factorized variational distribution $q_{\text{ent}}(\tau, g) = p(g) \Pi_{t=0}^{T-1} q_{\text{ent}}(a_t\mid s_t,g)$. The  variational distribution is defined by local distributions $q_{\text{ent}}(a\mid s, g) \coloneqq \pi_\theta(a\mid s,g) \exp(\hat{Q}^{\pi_\theta}(s,a,g)/\eta)$ for some temperature $\eta>0$ and estimates of Q-functions $\hat{Q}^{\pi_\theta}(s,a,g)$.
The design of $q_{\text{ent}}(\tau,g)$ could be interpreted as initializing $q_{\text{ent}}(a\mid s, g)$ with $ \pi_\theta(a\mid s,g)$ which effectively maximizes the \emph{second term} in \Cref{eq:e-step}, then taking one improvement step of the \emph{first term} \citep{abdolmaleki2018maximum}. This distribution satisfies \textbf{(P.1)} because the combined \gls{EM}-algorithm corresponds to  entropy-regularized policy iteration, and retains monotonic improvements in $J(\pi_\theta)$. However, it does not satisfy (\textbf{P.2}): when rewards are sparse $r(s,a,g)\approx 0$, estimates of Q-functions  are sparse $Q^{\pi_\theta}(s,a,g) \approx 0$ and leads to uninformed variational distributions $q_{\text{ent}}(a\mid s) \propto \pi_\theta(a\mid s)\exp(Q^{\pi_\theta}(s,a,g)/\eta) \approx \pi_\theta(a\mid s,g)$ for the M-step. In fact, when $\eta$ is large and the update to $q(a\mid s)$ from $\pi_\theta(a\mid s,g)$ becomes infinitesimal, the E-step is equivalent to policy gradients \citep{sutton1999,rauber2017hindsight}, which suffers from the sparsity of rewards as discussed in \Cref{label:hEM}.



\paragraph{Hindsight variational distribution.} Maximizing the \emph{first term} of the \gls{ELBO} is challenging when rewards are sparse. This motivates choosing a $q(\tau,g)$ which puts more weights on maximizing the \emph{first term}. Now, we formally introduce the hindsight variational distribution $q_h(\tau,g)$, the sampling distribution employed equivalently in \gls{HER} \citep{andrychowicz2017hindsight}. Sampling from this distribution is implicitly defined by an algorithmic procedure:
\begin{description}
    \item[Step 1.] Collect an on-policy trajectory or sample a trajectory from a replay buffer $\tau \sim \mathcal{D}$.
    \item[Step 2.]Find the $g$ such that the trajectory is rewarding, in that $R(\tau,g)$ is high or the trial is successful. Return the pair $(\tau,g)$.
\end{description}
Note that \textbf{Step 2} can be conveniently carried out with access to the reward function $r(s,a,g)$ as in \citep{andrychowicz2017hindsight}. Contrary to $q_{\text{ent}}(\tau,g)$, this hindsight variational distribution maximizes the \emph{first term} in \Cref{eq:e-step} by construction. This naturally satisfies (\textbf{P.2}) as $q_h(\tau,g)$ provides highly rewarding samples $(\tau,g)$ and hence dense signals to the M-step. The following theorem shows how $q_h(\tau,g)$ improves the sampling performance of our gradient estimates
\begin{theorem}
\label{thm:hindsight}
(Proof in \Cref{appendix:proof-hindsight}.) Consider the illustrative example in \Cref{thm:sparsity}. Let $\eta^h(a,g) = r(s,b,g^\prime) \nabla_{L_{a,g}} \log \pi(b\mid s,g^\prime) / k$ be the normalized one-sample \textsc{reinforce} gradient estimator where $(b,g^\prime)$ are sampled from the hindsight variational distribution with an on-policy buffer. Then the relative error  $\sqrt{\text{MSE}[\eta^h_{a,g}]} / \mathbb{E}[\eta_{a,g}] = \sqrt{k} (1 + o(1))$  grows sub-linearly for all $\forall a\in \mathcal{A},g\in \mathcal{G}$.
\end{theorem}
\Cref{thm:hindsight} implies that to further reduce the relative error of the hindsight estimator $\eta^h(a,g)$ with traditional Monte Carlo sampling would require $m \approx (\sqrt{k})^2 = k$ samples, which scales linearly with the problem size $k$. This is a sharp contrast to $m \approx k^2$ from using the on-policy \textsc{reinforce} gradient estimator. The above result shows the benefits of \gls{IS}, where under $q_h(\tau,g)$ rewarding trajectory-goal pairs are given high probabilities and this naturally alleviates the issue with sparse rewards. The following result shows that $q_h(\tau,g)$ also satisfies (\textbf{P.1}) under mild conditions.
\begin{theorem}
\label{thm:monotonic}
(Proof in \Cref{appendix:proof-mono}.)
Assume $p(g)$ to be uniform without loss of generality and a tabular representation of policy $\pi_\theta$. At iteration $t$, assume that the partial E-step returns $q_t(\tau,g)$ and the M-step objective in \Cref{eq:pi-update} is optimized fully. Also assume the variational distribution to be the hindsight variational distribution $q_t(\tau,g) \coloneqq q_h(\tau,g)$. Let $\tilde{p}_t(g) \coloneqq \int_\tau q_t(\tau,g) d\tau$ be the marginal distribution of goals. The performance is lower bounded as $J(\pi_{\theta_{t+1}}) \geq |\text{supp}(\tilde{p}_t(g))| / |\mathcal{G}| \eqqcolon \tilde{L}_t$. When the replay buffer size $\mathcal{D}$ increases over iterations, the lower bound improves $\tilde{L}_{t+1} \geq \tilde{L}_t$.
\end{theorem}

\subsection{Algorithm}
We now present \gls{hEM} which combines the above E- and M-steps. The algorithm maintains a policy $\pi_\theta(a\mid s,g)$. At each iteration, \gls{hEM} collects $N$ trajectory-goal pairs by first sampling a goal $g \sim p(\cdot)$ and then rolling out a trajectory $\tau$. Our initial experiments showed that exploration is critical when collecting trajectories. Under-exploration might drive \gls{hEM} to sub-optimal solutions. For continuous action space, we modify the agent to execute $a^\prime=\mathcal{N}(0,\sigma_a^2) + a$ where $a\sim\pi_\theta(a\mid s,g)$ and $\sigma_a=0.5$. This small amount of injected noise ensures that the algorithm has sufficient coverage of the state and goal space, which we find to be important for learning stability \citep{fortunato2017,plappert2018}. After this collection phase, all trajectories are stored into a replay buffer $\mathcal{D}$ \citep{mnih2013}.

At training time, \gls{hEM} carries out a partial E-step by sampling $(\tau,g)$ pairs from  $q_h(\tau,g)$. In practice, given a trajectory $\tau$ uniformly sampled from the buffer, a target goal $g$ is sampled using the \emph{future} strategy proposed in \citep{andrychowicz2017hindsight}. For the partial M-step, the policy is updated through stochastic gradient descents on \Cref{eq:pi-update} with the Adam optimizer \citep{kingma2014adam}. Importantly, \gls{hEM} is an off-policy \gls{RL} algorithm \emph{without} value functions, which also makes it agnostic to reward functions. The pseudocode is summarized in  \Cref{algo:routine}. Please refer to \Cref{appendix:experiment} for full descriptions of the algorithm.

\begin{algorithm}[!ht]
    \begin{algorithmic}[1]
        \STATE \textbf{INPUT} policy $\pi_\theta(a\mid s,g)$.
        \WHILE { $t=0,1,2...$ }
        \STATE Sample goal $g\sim p(\cdot)$ and trajectory $\tau \sim p(\cdot\mid\theta,g)$ by executing $\pi_\theta$ in the \gls{MDP}. Save data $(\tau,g)$ to a replay buffer $\mathcal{D}$.
        \STATE \textbf{E-step.} Sample from $q_h(\tau,g)$: sample $\tau \equiv (s_t,a_t)_{t=0}^{T-1} \sim \mathcal{D}$ and find rewarding goals $g$.
        \STATE \textbf{M-step.} Update the policy by a few gradient ascents  $\theta \leftarrow \theta + \nabla_\theta \log  \sum_{t=0}^{T-1} \log \pi_\theta(a_t\mid s_t,g)$.
        \ENDWHILE
    \end{algorithmic}
        \caption{Hindsight Expectation Maximization (\gls{hEM})}\label{algo:routine}
\end{algorithm}





\subsection{Connections to prior work}


\paragraph{Hindsight experience replay.} The core of \gls{HER} lies in the hindsight goal replay \citep{andrychowicz2017hindsight}. Similar to \gls{hEM}, \gls{HER} samples trajectory-goal pairs from the hindsight variational distribution $q_h(\tau,g)$ and minimize the Q-learning loss
$
    \mathbb{E}_{(\tau,g) \sim q_h(\cdot)} [ \sum_{t=0}^{T-1}  (Q_\theta(s_t,a_t,g) - r(s_t,a_t,g) - \gamma \max_{a^\prime} Q_\theta(s_i,a^\prime,g))^2]
$. The development of \gls{hEM} in \Cref{label:algo} formalizes this choice of the sampling distribution $q(\tau,g) \coloneqq q_h(\tau,g)$ as partially maximizing the \gls{ELBO} during an E-step. Compared to \gls{hEM}, \gls{HER} learns a critic $Q_\theta(s,a,g)$. We will see in the experiments that such critic learning tends to be much more unstable when rewards are sparse and inputs are high-dimensional, as was also observed in \citep{lee2019stochastic,kostrikov2020image}.

\paragraph{Hindsight policy gradient.} In its vanilla form, the \gls{HPG} considers on-policy stochastic gradient estimators of the \gls{RL} objective  \citep{rauber2017hindsight} as $
     \mathbb{E}_{p(g)p(\tau\mid\theta,g)} [ R(\tau,g) \nabla_\theta \log p(\tau\mid\theta,g)] $. Despite variance reduction methods such as  control variates \citep{sutton1999,rauber2017hindsight}, the unbiased estimators of \gls{HPG} do not address the rare event issue central to sparse rewards \gls{MDP}, where $R(\tau,g) \nabla_\theta \log p(\tau\mid\theta,g)$ taking non-zero values is a rare event under the on-policy measure $p(g)p(\tau\mid\theta,g)$. Contrast \gls{HPG} to the unbiased \gls{IS} objective in \Cref{eq:vae-rl-elbo}: $\mathbb{E}_{q(\tau,g)} [R(\tau,g) \nabla_\theta \log p(\tau\mid\theta,g) \cdot \frac{p(g)p(\tau\mid\theta,g)}{q(\tau,g)}]$, where the proposal $q(\tau,g)$ ideally prioritizes the rare events \citep{rubino2009rare} to generate rich learning signals. \gls{hEM} further avoids the explicit \gls{IS} ratios with the variational approach that leads to an \gls{ELBO} \citep{blei2017}.

\paragraph{Related work on learning from hindsight.} \citep{ding2019goal,ghosh2019learning} propose imitation learning algorithms for goal-conditioned \gls{RL}, which are similar to the M-step in Algorithm \ref{algo:routine}. While their algorithms are motivated from a purely behavior cloning perspective, we draw close connections between goal-conditioned \gls{RL} and probabilistic inference based on graphical models. This new perspective of \gls{hEM} decomposes the overall algorithms into two steps and clarifies their respective effects. Recently, \gls{hPI} \citep{eysenbach2020rewriting} propose to model the joint distribution over trajectories and goals $p(g,\tau)$, which leads to similar \gls{EM}-based updates as \gls{hEM}. Compared to \gls{hPI}, \gls{hEM} interprets the hindsight distribution (denoted as the \emph{relabeling} distribution in \citep{eysenbach2020rewriting}) as \gls{IS} proposals. In addition, \gls{hEM} is  specialized to sparse rewards while \gls{hPI} is designed and evaluated for generic goal-conditioned \gls{RL} problems with potentially dense reward signals.

With the E-step, \gls{hEM} assigns more weights to rewarding $(\tau,g)$ pairs. Beyond goal-conditioned \gls{RL}, This general idea of prioritizing samples with high returns has been combined with imitation learning \citep{oh2018self,guo2018generative,guo2019efficient}, Q-learning \citep{tang2020self} or model-based methods \citep{goyal2018recall}.

\paragraph{Supervised learning for \gls{RL}.} The idea of applying supervised learning techniques in an iterative \gls{RL} loop has been shown to stabilize the algorithms. In this space, successful examples include both model-based \citep{silver2016mastering,silver2018general,schrittwieser2020mastering,grill2020monte} and model-free algorithms \citep{levine2013guided,abdolmaleki2018maximum,vuong2018supervised,song2019v}. As also evidenced by our own experiments in the next section, supervised learning is especially helpful for problems with high-dimensional image-based inputs (see also, e.g., \citep{levine2016end,dosovitskiy2016learning,song2019v,schrittwieser2020mastering}).

\paragraph{Importance sampling in probabilistic inference.} Our work draws inspirations from the \gls{IS} views of recent probabilistic inference models. The use of \gls{IS} is inherent in the derivation of \gls{ELBO} \citep{blei2017}. Notably, for \gls{VAE} \citep{kingma2013auto},  \citep{burda2015importance} reinterpret the variational distribution $q_\phi(z\mid x)$ as an \gls{IS} proposal in place of the prior $p(z)$. This leads to a tighter \gls{ELBO} through the use of multiple importance-weighted samples, which is useful in some settings \citep{rainforth2018tighter}. We expect such recent developments in probabilistic inference literature to be useful for future work in goal-conditioned \gls{RL}.

\section{Experiments}
\label{sec:experiments}

We evaluate the empirical performance of \gls{hEM} on a wide range of goal-conditioned \gls{RL} benchmark tasks. These tasks all have extremely sparse binary rewards which indicate success of the trial.

\paragraph{Baselines.} Since \gls{hEM} builds on pure model-free concepts, we focus on the general purpose, model-free state-of-the-art algorithm \gls{HER} \citep{andrychowicz2017hindsight} as a comparison. In some cases we also compare with the closely-related \gls{HPG} \citep{rauber2017hindsight}. However, we find that even as \gls{HPG} adopts more dense rewards, its performance evaluated as the success rate is  inferior than \gls{HER} and \gls{hEM}. We do not compare with other model-free baselines such as HPI \citep{eysenbach2020rewriting}, because they adopt a formulation of dense rewards and their code is not publicly available, making it difficult to make meaningful comparison. We also do not compare with algorithms which assume structured knowledge about the task such as \citep{nair2018visual,nasiriany2019planning}, though their combination with \gls{hEM} is an interesting future direction. See \Cref{appendix:experiment} for more details.

\paragraph{Evaluations.} The evaluation criterion is the success rate at test time given a \emph{fixed} budget on the total number of samples collected during training. For all evaluations in \Cref{figure:intro}, \Cref{figure:robotics} and \Cref{figure:robotics-image}, we run $5$ runs of each algorithm and plot the $\text{mean}\pm\text{std}$ curves. Note that for many cases the standard deviations are small. We speculate this is because all experiments are run with $M\geq 20$ parallel workers for data collection and gradient computations,
which reduces the performance variance across seeds.

\paragraph{Implementation details.} See Appendix~\ref{appendix:experiment} for full details on parameterizations of the policy in discrete and continuous action space, network architecture, data collection and hyper-parameters on training.



\begin{figure*}[h]
\centering
\subcaptionbox{\textbf{Flip bit results}}[.24\linewidth]{\includegraphics[width=1.5in]{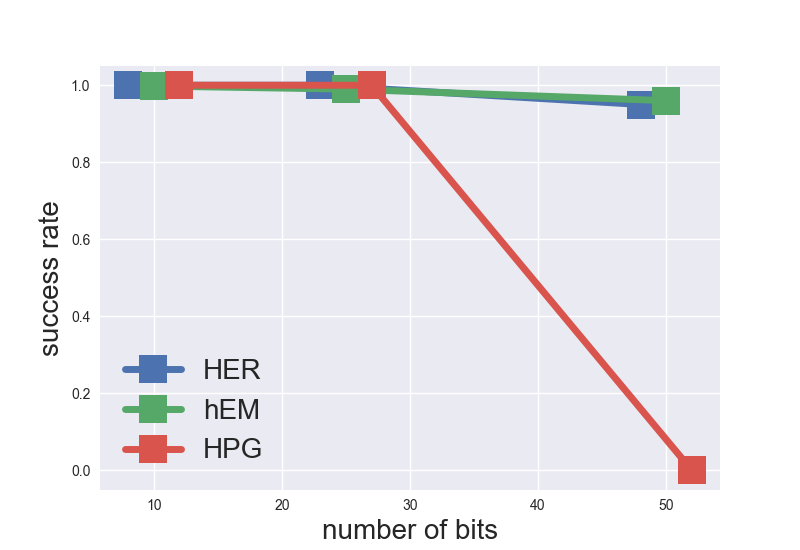}}
\subcaptionbox{\textbf{Flip bit curves}}[.24\linewidth]{\includegraphics[width=1.5in]{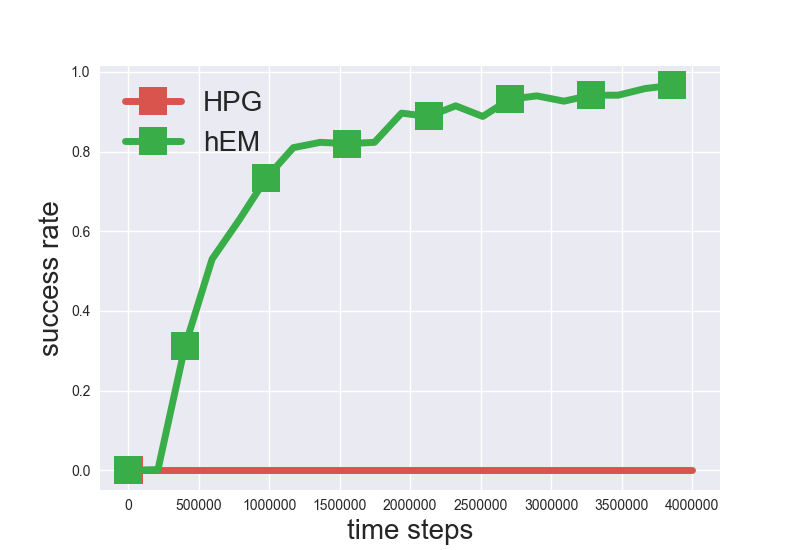}}
\subcaptionbox{\textbf{Navigation results}}[.24\linewidth]{\includegraphics[width=1.5in]{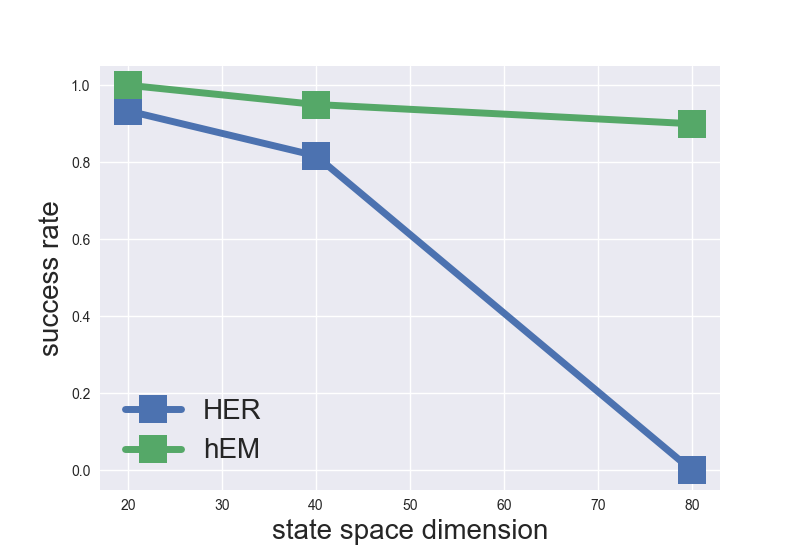}}
\subcaptionbox{\textbf{Navigation curves}}[.24\linewidth]{\includegraphics[width=1.5in]{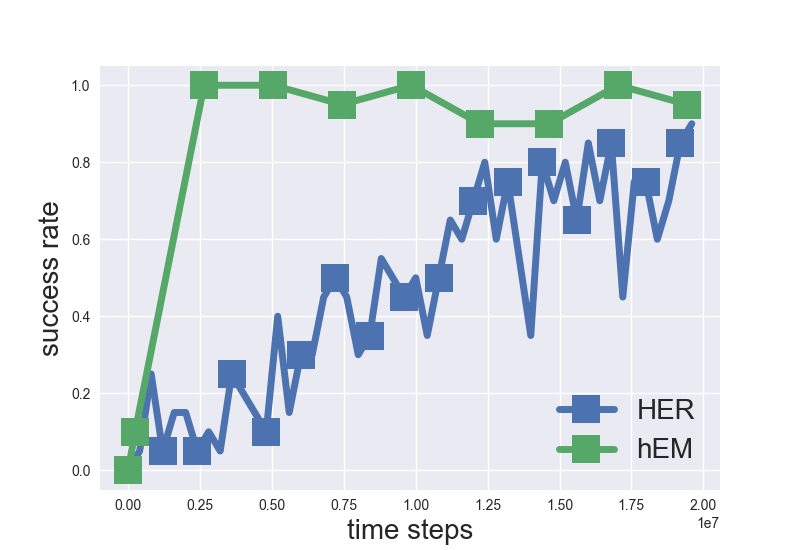}}
\caption{\small{Summary of results for Flip bit and continuous navigation \gls{MDP}. Plots (a) and (c) show the final performance after training is completed. Plots (b) and (d) show the training curves for Flip bit $K=50$ and navigation $K=40$ respectively. \gls{hEM} consistently outperforms \gls{HER} and \gls{HPG} across these tasks.}}
\label{figure:flipbit}
\end{figure*}

\subsection{Simple examples}

\paragraph{Flip bit.} Taken from \citep{andrychowicz2017hindsight}, the \gls{MDP} is parameterized by the number of bits $K$. The state space and goal space $\mathcal{S} = \mathcal{G} = \{0,1\}^K$ and the action space $\mathcal{A} = \{1,2, \ldots, K\}$. Given $s_t$, the action flips the bit at location $a_t$. The reward function is $r(s_t,a_t) = \mathbb{I}[s_{t+1}=g]$
, the state is flipped to match the target bit string. The environment is difficult for traditional \gls{RL} methods as the search space is of exponential size $|\mathcal{S}|=2^K$. In \Cref{figure:flipbit}(a), we present results for \gls{HER} (taken from Figure 1 of \citep{andrychowicz2017hindsight}), \gls{hEM} and \gls{HPG}. Observe that \gls{hEM} and \gls{HER} consistently perform well even when $K=50$ while the performance of \gls{HPG} drops drastically as the underlying spaces become enormous. See \Cref{figure:flipbit}(b) for the training curves of \gls{hEM} and \gls{HPG} for $K=50$; note that \gls{HPG} does not make any progress.

\paragraph{Continuous navigation.} As a continuous analogue of the Flip bit \gls{MDP}, consider a $K$-dimensional navigation task with a point mass. The state space and goal space coincide with $\mathcal{X} = \mathcal{G} = [-1,1]^K$ while the actions $\mathcal{A}=[-0.2,0.2]^K$ specify changes in states. The reward function is $r(s_t,a_t) = \mathbb{I}[\;\|\;s_{t+1}-g\;\|\; < 0.1]$ which indicates success when reaching the goal location. Results are shown in \Cref{figure:flipbit}(c) where we see that as $K$ increases, the search space quickly explodes and the performance of \gls{HER} degrades drastically. The performance of \gls{hEM} is not greatly influenced by increases in $K$. See \Cref{figure:flipbit}(d) for the comparison of training curves between \gls{hEM} and \gls{HER} for $K=40$. \gls{HER} already learns much more slowly compared to \gls{hEM} and degrades further when $K=80$.

\subsection{Goal-contitioned reaching tasks}
\label{sub:goal-conditioned-tasks}

To assess the performance of \gls{hEM} in contexts with richer transition dynamics, we consider a wide range of goal-conditioned reaching tasks. We present details of their state space $\mathcal{X}$, goal space $\mathcal{G}$ and action space $\mathcal{A}$ in \Cref{appendix:experiment}. These include  \textbf{Point mass}, \textbf{Reacher goal}, \textbf{Fetch robot} and \textbf{Sawyer robot}, as illustrated in \Cref{figure:tasks} in \Cref{appendix:experiment}.

Across all tasks, the reward takes the sparse form $r(s,a,g) = \mathbb{I}[\text{success}]$. As a comparison, we also include a \gls{HER} baseline where the rewards take the form $\tilde{r}(s,a,g) = -\mathbb{I}[\text{failure}]$. Such reward shaping does not change the optimality of policies as $\tilde{r} = r - 1$ and is also the default rewards employed by e.g., the Fetch robot tasks. Surprisingly, this  transformation has a big impact on the performance of \gls{HER}. We denote the \gls{HER} baseline under the reward $r=\mathbb{I}[\text{success}]$ as `\gls{HER}(0/1)' and $\tilde{r} = -\mathbb{I}[\text{failure}]$ as `\gls{HER}(-1/0)'.

\begin{figure*}[h]
\centering
\subcaptionbox{\textbf{Point mass}}[.24\linewidth]{\includegraphics[width=1.5in]{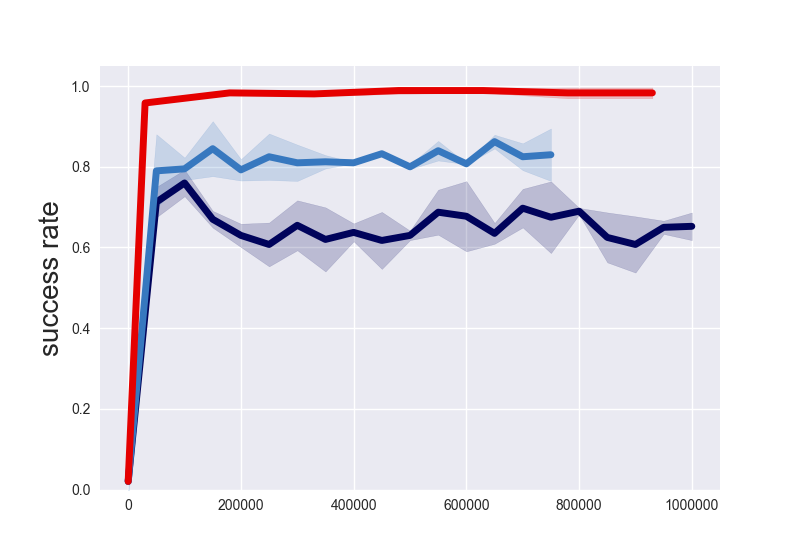}}
\subcaptionbox{\textbf{Reacher goal}}[.24\linewidth]{\includegraphics[width=1.5in]{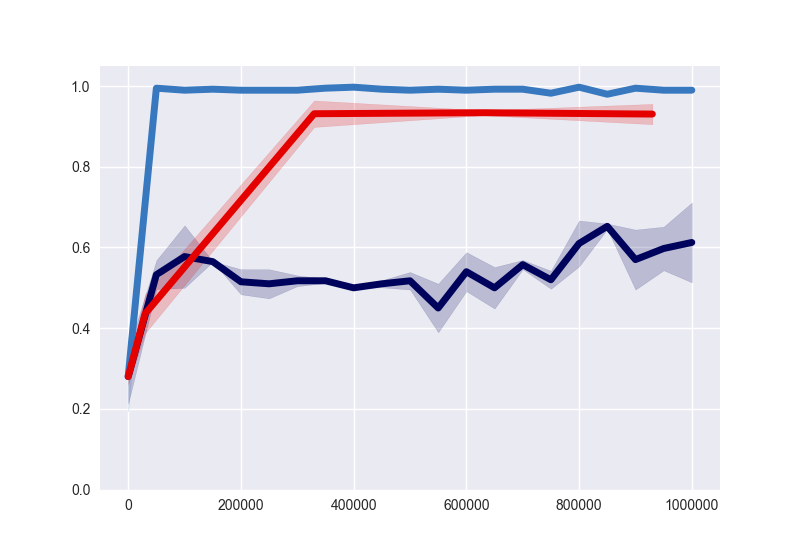}}
\subcaptionbox{\textbf{Fetch robot}}[.24\linewidth]{\includegraphics[width=1.5in]{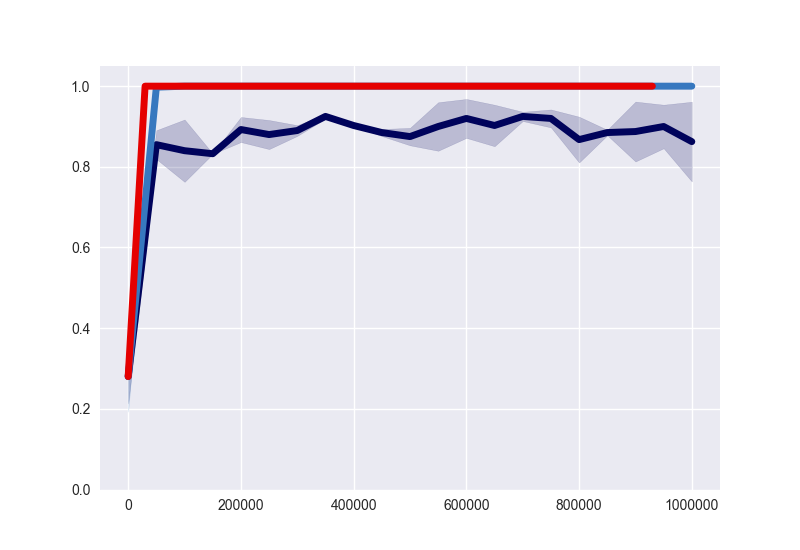}}
\subcaptionbox{\textbf{Sawyer robot}}[.24\linewidth]{\includegraphics[width=1.5in]{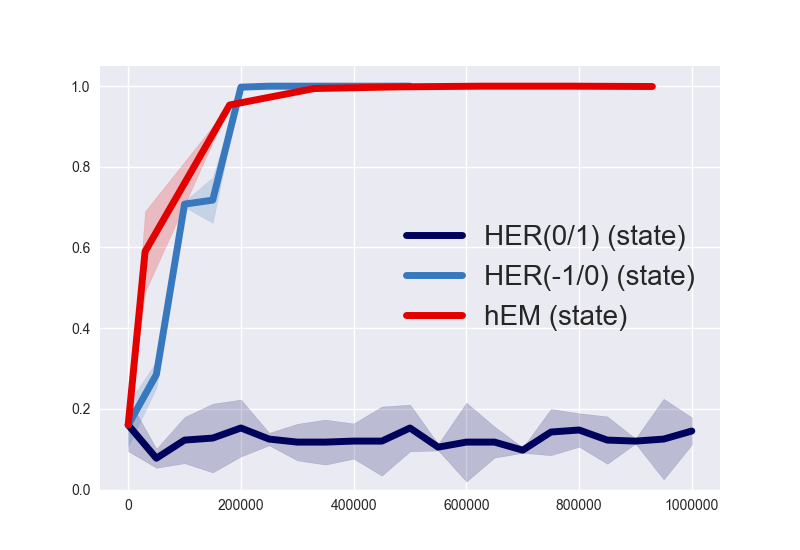}}
\caption{\small{Training curves of \gls{hEM} and \gls{HER} on four goal-conditioned \gls{RL} benchmark tasks with state-based inputs and sparse binary rewards. The y-axis shows the success rates and the x-axis shows the training time steps. All curves are calculated based on averages over $5$ random seeds.  Standard deviations are small across seeds.}}
\label{figure:robotics}
\end{figure*}

 From the result in \Cref{figure:robotics}, we see that \gls{hEM} performs significantly better than \gls{HER} with binary rewards (\gls{HER}-sparse). The performance of \gls{hEM} quickly converges to optimality while \gls{HER} struggles at learning good Q-functions. However, when compared with \gls{HER}($-1/0$), \gls{hEM} does not achieve noticeable gains. Such an observation confirms that \gls{HER} is sensitive to reward shaping due to the critic learning.


\begin{figure*}[h]
\centering
\subcaptionbox{\textbf{Point mass (I)}}[.24\linewidth]{\includegraphics[width=1.5in]{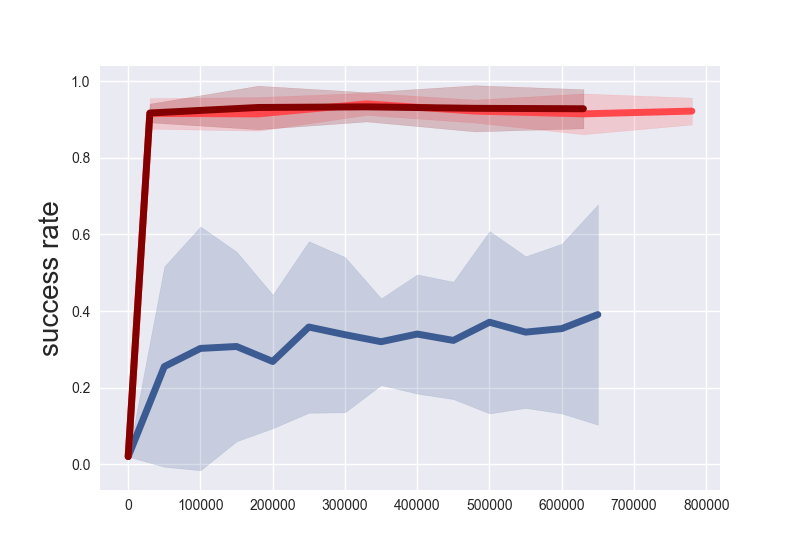}}
\subcaptionbox{\textbf{Reacher goal (I)}}[.24\linewidth]{\includegraphics[width=1.5in]{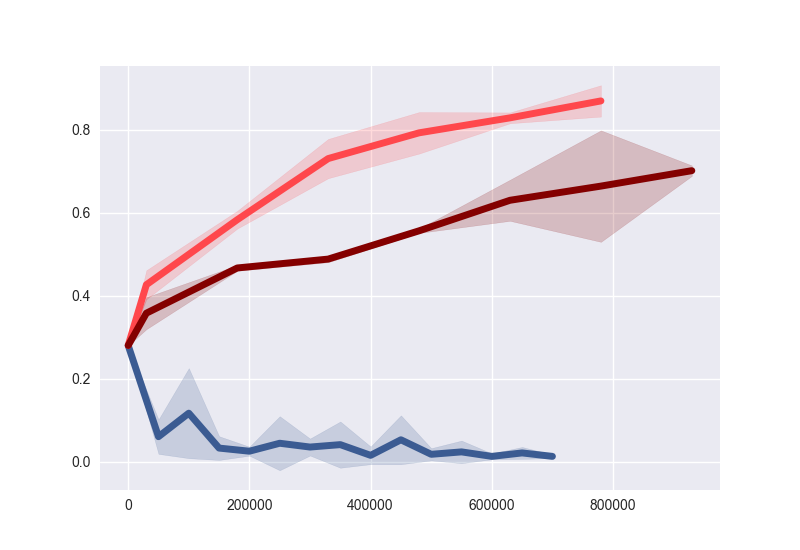}}
\subcaptionbox{\textbf{Fetch robot (I)}}[.24\linewidth]{\includegraphics[width=1.5in]{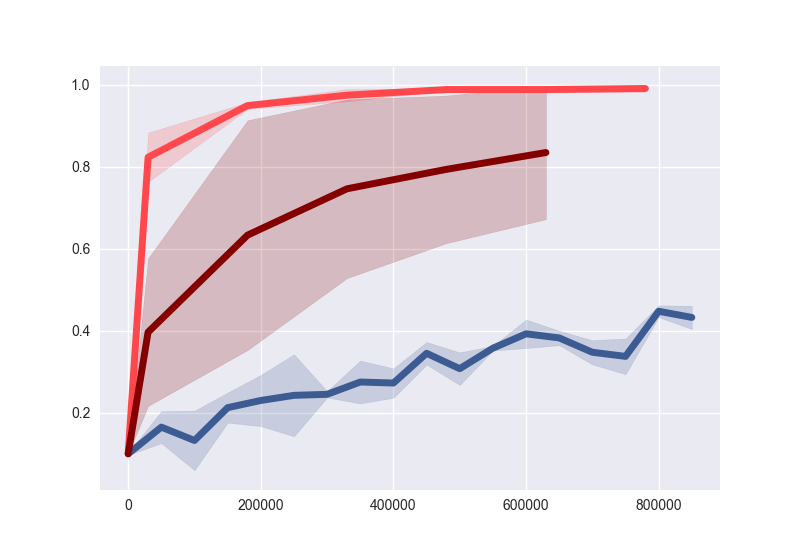}}
\subcaptionbox{\textbf{Sawyer robot (I)}}[.24\linewidth]{\includegraphics[width=1.5in]{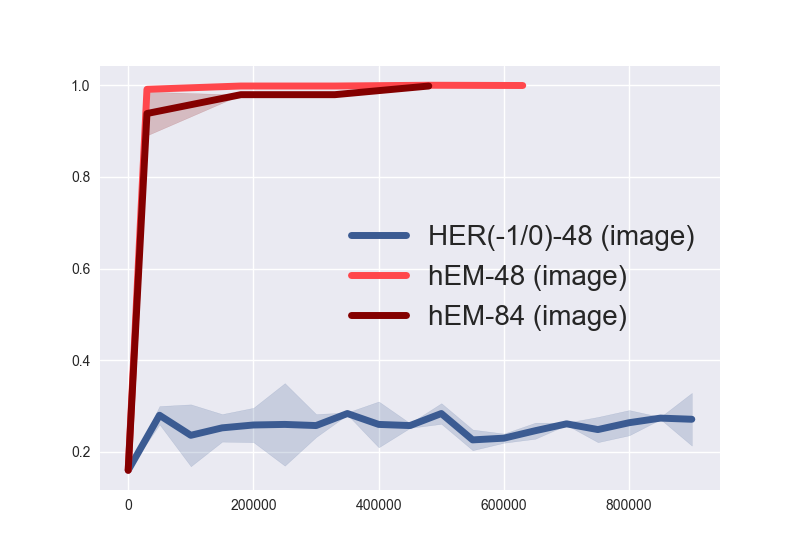}}
\caption{\small{Training curves of \gls{hEM} and \gls{HER} on four goal-conditioned \gls{RL} tasks with image-based inputs. Standard deviations are small across seeds. All curves are calculated based on averages over $5$ random seeds.  `\gls{hEM}-48' refers to image inputs with $w=48$. \gls{hEM} achieves stable learning regardless of the input sizes, though larger sizes in general slow down the learning speed.}}
\label{figure:robotics-image}
\end{figure*}

\subsection{Image-based tasks}
We further assess the performance of \gls{hEM} when policy inputs are high-dimensional images (see \Cref{figure:tasks-images} for illustrations). Across all tasks, the state inputs are by default images $s \in \mathbb{R}^{w\times w\times 3}$ where $w\in\{48,84\}$ while the goal is still low-dimensional. See \Cref{appendix:experiment} for the network architectures.


We focus on the comparison between \gls{hEM} and \gls{HER}($-1/0$) in \Cref{figure:robotics-image}, as the performance of \gls{HER} with binary rewards is inferior as seen \Cref{sub:goal-conditioned-tasks}. We see that for image-based tasks,  \gls{HER}($-1/0$) significantly underperforms \gls{hEM}. While \gls{HER}($-1/0$) makes slow progress for most cases, \gls{hEM} achieves stable learning across all tasks. We speculate this is partly due to the common observations \citep{lee2019stochastic,kostrikov2020image} that \gls{TD}-learning  directly from high-dimensional image inputs is challenging. For example, prior work \citep{nair2018visual} has applied a \gls{VAE}  \citep{kingma2013auto} to reduce the dimension of the image inputs for downstream \gls{TD}-learning. On the contrary, \gls{hEM} only requires optimization in a supervised learning style, which is much more stable with end-to-end training on image inputs.

Further, image-based goals are much easier to specify in certain contexts \citep{nair2018visual}. We evaluate \gls{hEM} on image-based goals for the Sawyer robot and achieve similar performance as the state-based goals. See results in \Cref{figure:ablation-appendix} in \Cref{appendix:experiment}.

\subsection{Goal-conditioned Fetch tasks}

Finally, we evaluate the performance of \gls{HER} and \gls{hEM} over the full suite of Fetch robot tasks introduced in \citep{andrychowicz2017hindsight}. These Fetch tasks have the same transition dynamics as the previous Fetch Reach task, but differ significantly in the goal space. As a result, some of the tasks are more challenging than Fetch Reach due to difficulties in efficiently exploring the goal space. To tackle this issue, we implement a hybrid algorithm between \gls{hEM} and \gls{HER} by combining their policy updates. In particular, we propose to update policy $\pi_\theta$ by considering the following hybrid objective,
\begin{align*}
    L(\theta) = &\underbrace{\mathbb{E}_{(s,g)\sim\mathcal{D}} \left[Q_\phi \left(s,\pi_\theta(s),g\right)\right]}_{\text{HER}} + \\ &\eta \cdot \underbrace{\mathbb{E}_{q(\tau,g)}\left[\sum_{t=0}^{T-1} \log \pi_\theta(a_t \mid s_t,g) \right]}_{\text{hEM}}.
\end{align*}
The policy is updated as $\theta\leftarrow \theta + \nabla_\theta L(\theta)$. The Q-function $Q_\phi(s,a,g)$ is trained with \gls{TD}-learning to approximate the true Q-function $Q_\phi\approx Q^{\pi_\theta}$. We see that the two terms echo the  loss functions employed by \gls{HER} and \gls{hEM} respectively. The constant coefficient $\eta\geq 0$ trades-off the two loss functions. We find $\eta=0.1$ to perform uniformly well on all selected tasks.

Besides loss functions, the algorithm collects data using the same procedure as \gls{HER}.
This allows \gls{hEM} to leverage potentially more efficient exploration schemes of \gls{HER} (e.g., correlated exploration noises). We find that the hybrid algorithm achieves marginal improvements in the learning speed compared to \gls{HER}, see \Cref{figure:ablation-paper} for the comparison and full results in Appendix \ref{appendix:experiment}.

\paragraph{Remark.} Exploration is an under-explored yet critical area in goal-conditioned \gls{RL}, especially for tasks with hierarchical goal space and long horizons. Consistent with the observations in recent work \citep{andrychowicz2017hindsight}, we find that under-exploration could lead to sub-optimal policies. We believe when combined with recent advances in exploration for goal-conditioned \gls{RL} \citep{zhang2020automatic,pitis2020maximum}, the performance of \gls{hEM} could be further stabilized.

\begin{figure}[h]
\centering
\subcaptionbox{\textbf{ Push}}[.48\linewidth]{\includegraphics[width=1.5in]{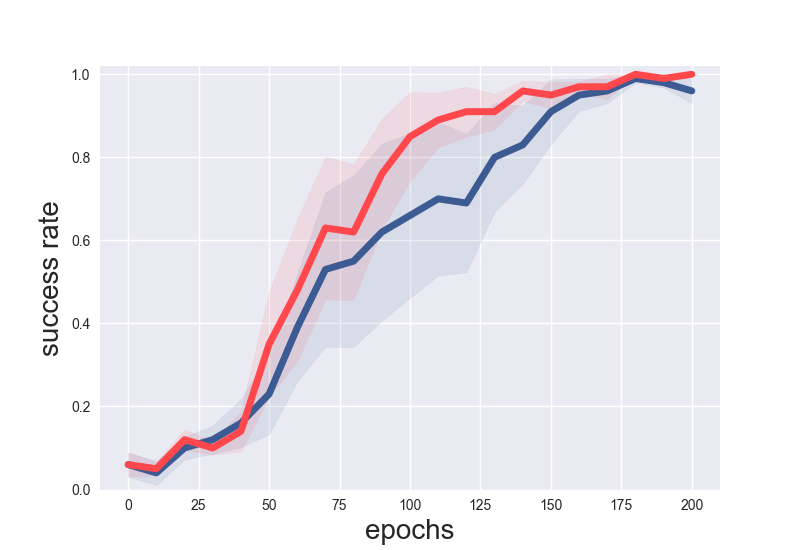}}
\subcaptionbox{\textbf{Pick and Place}}[.48\linewidth]{\includegraphics[width=1.5in]{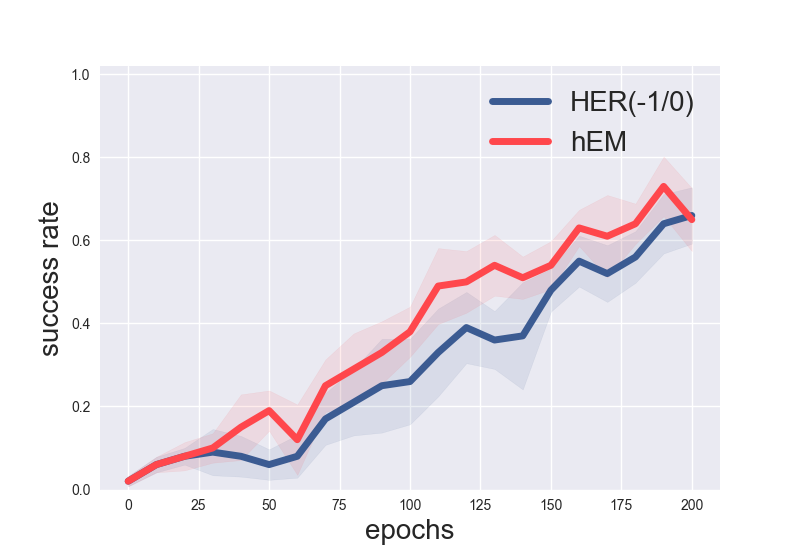}}
\caption{\small{Fetch tasks. Training curves of \gls{hEM} and \gls{HER} on two standard Fetch tasks. \gls{hEM} provides marginal speed up compared to \gls{HER}. All curves are calculated based on $5$ random seeds. The x-axis shows the training epochs.}}
\label{figure:ablation-paper}
\end{figure}

\subsection{Ablation study}

 Recall that the \gls{hEM} alternates between collecting $N$ trajectories and updating parameters via the \gls{EM} algorithm. We find the hyper-parameter $N$ impacts the algorithm significantly. Intuitively, the amount of collected data at each iteration implicitly determines the effective coverage of the goal space through exploration. When $N$ is small, the training might easily converge to local optima. See  \Cref{figure:ablation-appendix} in \Cref{appendix:experiment} for full results. Both tasks in the ablation are challenging, due to an enormous state space or the high dimensionality of the inputs. In general, we find that larger $N$ leads to better performance (note that here the performance is always measured with a fixed budget on the total time steps). Similar observations have been made for \gls{HER} \citep{andrychowicz2017hindsight}, where increasing the number of parallel workers generally improves training performance.

\section{Conclusion}

We present a probabilistic framework for goal-conditioned \gls{RL}. This framework motivates the development of \gls{hEM}, a simple and effective off-policy \gls{RL} algorithm. Our formulation draws formal connections between hindsight goal replay \citep{andrychowicz2017hindsight} and \gls{IS} for rare event simulation. \gls{hEM} combines the stability of supervised learning updates via the M-step and the hindsight replay technique via the E-step. We show improvements over a variety of benchmark \gls{RL} tasks, especially in high-dimensional input settings with sparse binary rewards.

\paragraph{Acknowledgements.} The authors would like to acknowledge the computation support of Google Cloud Platform.

\bibliographystyle{unsrt}
\bibliography{BIB}

\onecolumn
\appendix

\section{Details on Graphical Models for Reinforcement Learning}
\label{appendix:graphicalmodels}

In this section, we review details of the \emph{\gls{RL} as inference} framework \citep{haarnoja2018soft,levine2018reinforcement} and highlight its critical differences from \emph{Variational \gls{RL}}.

\begin{figure}[h]
\centering
\subcaptionbox{\small{Probabilistic inference}}[.25\linewidth]{
\begin{tikzpicture}
	\begin{pgfonlayer}{nodelayer}
		\node [style=filled] (0) at (0, 0) {$x$};
		\node [style=hollow] (1) at (0, 1.5) {$z$};
		\node [style=box] (4) at (-1.2, 1.5) {$\theta$};
		\node [style=box] (5) at (1.2, 1.5) {$\phi$};
		\node [style=empty] (10) at (0, 0.66) {};
		\node [style=empty] (11) at (-0.5, 0.66) {};
		\node [style=empty] (12) at (0.5, 0.66) {};
	\end{pgfonlayer}
	\begin{pgfonlayer}{edgelayer}
		\draw [style=arrow] (1) to (0);
		\draw [style=arrow] (4) to (1);
		\draw [style=arrow][bend right=60,dashed] (0) to (1);
		\draw [style=arrow][dashed] (5) to (1);
	\end{pgfonlayer};
\end{tikzpicture}
}
\subcaptionbox{\small{Variational \textsc{rl}}}[.23\linewidth]{
\begin{tikzpicture}
	\begin{pgfonlayer}{nodelayer}
		\node [style=filled] (0) at (0, 0) {$O$};
		\node [style=hollow] (1) at (0, 1.5) {$\tau$};
		\node [style=box] (4) at (-1.2, 1.5) {$\theta$};
		\node [style=box] (5) at (1.2, 1.5) {$q$};
		\node [style=empty] (10) at (0, 0.66) {};
		\node [style=empty] (11) at (-0.5, 0.66) {};
		\node [style=empty] (12) at (0.5, 0.66) {};
	\end{pgfonlayer}
	\begin{pgfonlayer}{edgelayer}
		\draw [style=arrow] (1) to (0);
		\draw [style=arrow] (4) to (1);
		\draw [style=arrow][dashed] (5) to (1);
	\end{pgfonlayer};
\end{tikzpicture}
}
\subcaptionbox{\small{\gls{RL} as inference}}[.23\linewidth]{
\begin{tikzpicture}
	\begin{pgfonlayer}{nodelayer}
		\node [style=filled] (0) at (0, 0) {$O$};
		\node [style=hollow] (1) at (0, 1.5) {$\tau$};
		\node [style=box] (5) at (1.2, 1.5) {$\theta$};
		\node [style=empty] (10) at (0, 0.66) {};
		\node [style=empty] (11) at (-0.5, 0.66) {};
		\node [style=empty] (12) at (0.5, 0.66) {};
	\end{pgfonlayer}
	\begin{pgfonlayer}{edgelayer}
		\draw [style=arrow] (1) to (0);
		\draw [style=arrow][dashed] (5) to (1);
	\end{pgfonlayer};
\end{tikzpicture}
}
\caption{\small{Plot (c) shows the graphical model for \emph{\gls{RL} as inference} \citep{haarnoja2018soft,levine2018reinforcement}. Solid lines represent generative models and dashed lines represent inference models. Circles represent random variables and squares represent parameters. Filled circles represent observed random variables. This graphical model does not have trainable parameters for the generative model. The policy dependent parameter $\theta$ is in the inference model.}}
\label{figure:appendix-rlasinference}
\end{figure}
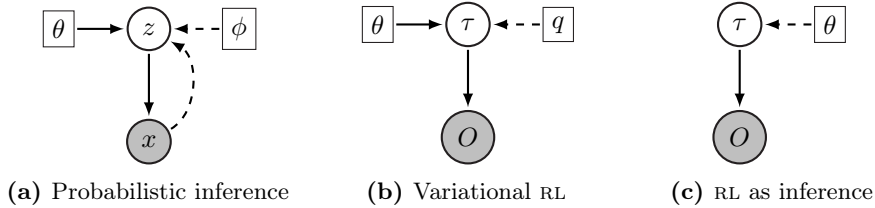

The graphical model for \emph{\gls{RL} as inference} is shown in \Cref{figure:appendix-rlasinference}(c). The framework also assumes a trajectory variable $\tau \equiv (s_t,a_t)_{t=0}^{T-1}$ which encompasses the state and action sequences. Conditional on the trajectory variable $\tau$, the optimality variable is defined as $p(O=1\mid\tau) \propto \exp (\sum_{t=0}^{T-1} r(s_t,a_t) / \alpha)$ for $\alpha > 0$. Under this framework, the trajectory variable has a prior $a_t \sim p(\cdot)$ where $p(\cdot)$ is usually set to be a uniform distribution over the action space $\mathcal{A}$.

The policy parameter $\theta$ comes into play with the inference model. The framework asks the question: what is the posterior distribution $p(\tau\mid O=1)$. To approximate this intractable posterior distribution, consider the variational distribution $q(\tau) \coloneqq \Pi_{t=0}^{T-1} \pi_\theta(a_t\mid s_t) p(s_{t+1}\mid s_t,a_t)$. Searching for the best approximate posterior by minimizing the KL-divergence $\mathbb{KL}[q(\tau)\;\|\;p(\tau\mid O=1)]$, it can be shown that this is equivalent to maximum-entropy \gls{RL} \citep{ziebart2008maximum,fox2015taming,asadi2017alternative}. It is important to note that \emph{\gls{RL} as inference} does not contain trainable parameters for the generative model.

Contrasting this to \emph{Variational \gls{RL}} and the graphical model for goal-conditioned \gls{RL} in \Cref{figure:rlasinference}: the policy dependent parameter $\theta$ is part of a generative model. The variational distribution $q(\tau,g)$, defined separately from $\theta$, is the inference model. In such cases, the variational distribution $q(\tau,g)$ is an auxiliary distribution which aids in the optimization of $\theta$ by performing partial E-steps.

\section{Details on proof}
\label{appendix:proof-quiv}
\subsection{Proof of Proposition \Cref{prop:equiv}}
The proof follows from the observation that $p(O=1) = \mathbb{E}_{g\sim p(\cdot),\pi}[p(O=1\mid\tau,g)] = J(\pi_\theta)$, and taking the log does not change the optimal solution.

\subsection{Proof of \Cref{thm:sparsity}}
\label{appendix:proof-sparsity}
Recall that we have a one-step \gls{MDP} setup where $\mathcal{A} = \mathcal{G}$ and $|\mathcal{A}| = k$. The policy $\pi(a\mid s,g) = \text{softmax}(L_{a,g})$ is parameterized by logits $L_{a,g}$. When the policy is initialized randomly, we have $L_{a,g}\equiv L$ for some $L$ and $\pi(a\mid s,g) = 1/k$ for all $a,g$. Assume also $p(g) = 1/k,\forall g$.

The one-sample \textsc{reinforce} gradient estimator for the component $L_{a,g}$ is $\eta_{a,g} = r(s,b,g^\prime) \log_{L_{a,g}} \pi(b\mid s,g^\prime)$ with $g^\prime \sim p(\cdot)$ and $b \sim \pi(\cdot\mid s,g^\prime)$. Further, we can show
\begin{align*}
     \mathbb{E}[\eta_{a,g}] = \frac{1}{k^2} \delta_{a,g} - \frac{1}{k^3},\ \mathbb{V}[\eta_{a,g}] = (\frac{1}{k^2} + \frac{2}{k^5} - \frac{2}{k^3} - \frac{1}{k^4}) \delta_{a,g} + \frac{1}{k^4} - \frac{1}{k^6},
 \end{align*}
 where $\delta_{a,b}$ are dirac-delta functions, which mean $\delta_{a,b}=1$ if $a=b$ and $\delta_{a,b}=0$ otherwise. Taking the ratio, we have the squared relative error (note that the estimator is unbiased and MSE consists purely of the variance)
 \begin{align*}
     \frac{\text{MSE}[\eta_{a,g}]}{\mathbb{E}[\eta_{a,g}]^2} = \frac{(k^4 + o(k^4)) \delta_{a,g} + (k^2 + o(k^2))}{(k^2 + o(k^2)) \delta_{a,g} + 1}
 \end{align*}
 The expression takes different forms based on the delta-function $\delta_{a,g}$. However, in either case (either $\delta_{a,g}=1$ or $\delta_{a,g} = 0$), it is clear that $\frac{\text{MSE}[\eta_{a,g}]}{\mathbb{E}[\eta_{a,g}]^2} = k^2 (1 + o(1))$, which directly reduces to the result of the theorem.

 \paragraph{Comment on the control variates.} We also briefly study the effect of control variates. Let $X,Y$ be two random variables and assume $\mathbb{E}[Y] = 0$. Then compare the variance of $\mathbb{V}[X]$ and $\mathbb{V}[X+\alpha Y]$ where $\alpha$ is chosen optimally to minimize the variance of the second estimator. It can be shown that with the best $\alpha^\ast$, the ratio of variance reduction is $(\mathbb{V}[X] - \mathbb{V}[X+\alpha^\ast Y]) / \mathbb{V}[X] = \rho^2 \coloneqq \text{Cov}^2[X,Y] / \mathbb{V}[X]\mathbb{V}[Y]$. Consider the state-based control variate for the \textsc{reinforce} gradient estimator, in this case $-\alpha \cdot \nabla_{L_{a,g}} \pi(b\mid s,g^\prime)$ where $\alpha$ is chosen to minimize the variance of the following aggregate estimator
 \begin{align*}
     \eta_{a,g}(\alpha)= r(s,b,g^\prime) \log_{L_{a,g}} \pi(b\mid s,g^\prime) - \alpha  \log_{L_{a,g}} \pi(b\mid s,g^\prime).
 \end{align*}
 Note that in practice, $\alpha$ is chosen to be state-dependent for \textsc{reinforce} gradient estimator of general MDPs and is set to be the value function $\alpha \coloneqq V^\pi(s)$. Such a choice is not optimal \citep{fu2015stochastic} but is conveniently adopted in practice. Here, we consider an optimal $\alpha^\ast$ for the one-step \gls{MDP}. The central quantity is the squared correlation $\rho^2$ between $r(s,b,g^\prime) \log_{L_{a,g}} \pi(b\mid s,g^\prime)$ and $\log_{L_{a,g}} \pi(b\mid s,g^\prime)$. With similar computations as above, it can be shown that
 $\rho^2 \approx 1$ for $b\neq g^\prime$ and $\rho^2 \approx \frac{1}{k^2}$ otherwise. This implies that for $k$ out of $k^2$ logits parameters, the variance reduction is significant; yet for the rest of the $k^2-k$ parameters, the variance reduction is negligible. Overall, the analysis reflects that conventional control variantes do not address the issue of sup-linear growth of relative errors as a result of \emph{sparse gradients}.

 \subsection{Proof of \Cref{thm:hindsight}}
 \label{appendix:proof-hindsight}

 The key to the proof is the same as the proof of \cref{thm:sparsity}: we analytically compute $\mathbb{E}[\eta_{a,g}]$ and $\mathbb{V}[\eta_{a,g}]$. We can show
 \begin{align*}
     \mathbb{E}[\eta_{a,g}] = \frac{1}{k^2} \delta_{a,g} + o(\frac{1}{k^2}),\ \mathbb{V}[\eta_{a,g}] = \frac{1}{k^3}\delta_{a,g} + o(\frac{1}{k^3}),
 \end{align*}

 More specifically, it is possible to show that the normalized one-step \textsc{reinforce} gradient estimator $\eta^h_{a,g} = r(s,b,g^\prime) \nabla_{L_{a,g}} \log \pi(b\mid s,g^\prime) / k$ with $(b,g^\prime) \sim q_h(\tau,g)$ has the following property
 \begin{align*}
     \frac{\text{MSE}[\eta_{a,g}]}{\mathbb{E}[\eta_{a,g}]^2} = \frac{(k^3 + o(k^3)) \delta_{a,g} + (k + o(k))}{(k^2 + o(k)) \delta_{a,g} + 1}.
 \end{align*}
 The above equality implies the result of the theorem. Indeed,  the above implies regardless of whether $\delta_{a,g}=0$ or $\delta_{a,g}=1$: $ \frac{\text{MSE}[
 \eta_{a,g}]}{\mathbb{E}[\eta_{a,g}]^2} = k(1+o(1))$.

 \paragraph{Remark.} Contrast this with the result from \Cref{thm:sparsity}, where the result is $k^2(1+o(1))$. The main difference stems from the variance: in \Cref{thm:sparsity} the variance is of order $\frac{1}{k^2}$ while here the variance is of order $\frac{1}{k^3}$. The variance reduction leads to significant improvements of the sample efficiency of the estimation.

 \subsection{Proof of \Cref{thm:monotonic}}
 \label{appendix:proof-mono}
 Without loss of generality we assume $p(g)$ is a uniform measure, i.e. $p(g) = 1/|\mathcal{G}|$. If not, we could always find a transformation $g = f(g^\prime)$ such that $g^\prime$ takes a uniform measure \citep{casella2002statistical} and treat $g^\prime$ as the goal to condition on.

 Let $|\mathcal{G}| < \infty$ and recall $\text{supp}(\tilde{p}(g))$ to be the support of $\tilde{p}(g)$. The uniform distribution assumption deduces that $|\text{supp}(\tilde{p}(g))| = \int_{g\in \text{supp}(\tilde{p}(g))} dg$. At iteration $t$, under tabular representation, the M-step update implies that $\pi_\theta$ learns the optimal policy for all $g$ that could be sampled from $q(\tau,g)$, whhich effectively corresponds to the support of $\tilde{p}(g)$. Formally, this implies $\mathbb{E}_{p(\tau\mid \theta_{t+1},g)}[R(\tau,g)] = 1$ for $\forall g \in \text{supp}(\tilde{p}_t(g))$. This further implies
 \begin{align}
     J(\pi_{\theta_{t+1}}) \coloneqq \int \mathbb{E}_{p(\tau\mid \theta_{t+1},g)}[R(\tau,g)] p(g) dg \geq \int_{g\in \text{supp}(\tilde{p}_t(g))}  1 \cdot p(g) dg
     = |\text{supp}(\tilde{p}_t(g))| / |\mathcal{G}|. \nonumber
 \end{align}

 \section{Additional Experiment Results}
 \label{appendix:experiment}

 \subsection{Details on Benchmark tasks}
 All reaching tasks are built with physics simulation engine MuJoCo \citep{todorov2012}. We build customized point mass environment;
 the Reacher and Fetch environment is partly based on OpenAI gym environment \citep{brockman2016}; the Sawyer environment is based on the multiworld open source code \url{https://github.com/vitchyr/multiworld}.

\begin{figure}[h]
\centering
\includegraphics[width=6in]{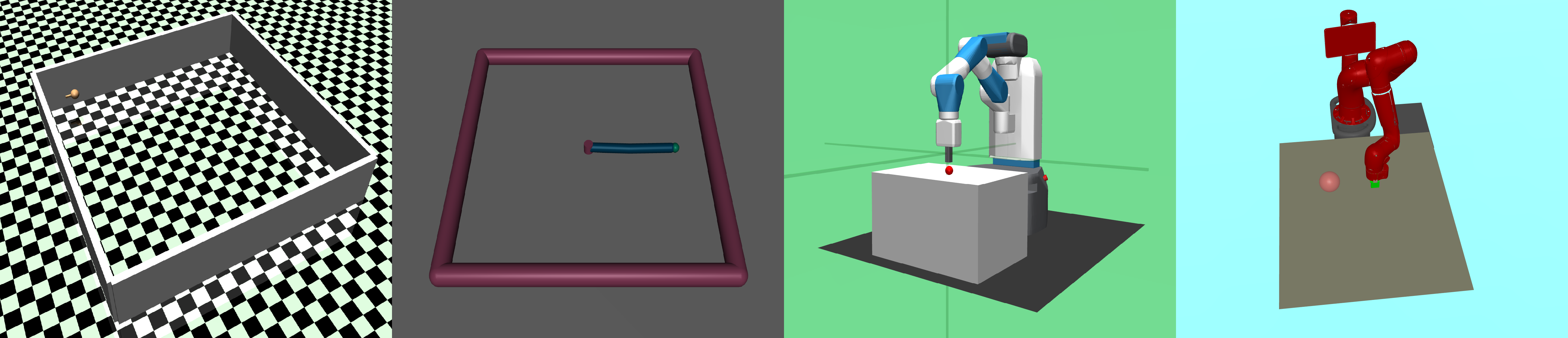}
 \caption{\small{Illustration of tasks. From left to right: Point mass, Reacher, Fetch robot and Sawyer Robot. On the right is the image-based input for Fetch robot. For additional information on the tasks, see \Cref{appendix:experiment}.}}
\label{figure:tasks}
\end{figure}

\begin{figure}[h]
\centering
\includegraphics[width=6in]{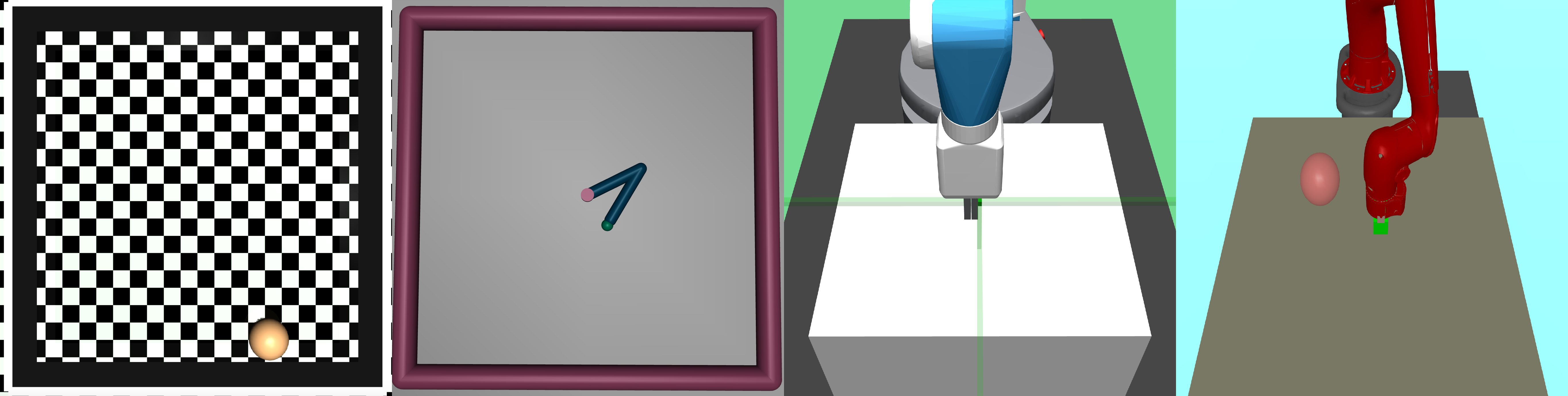}
  \caption{\small{Illustration of image-based inputs for different reaching tasks in the main paper. Images are down-sampled to be of size $w \times w \times 3$ as inputs, where $w\in \{48,84\}$.}}
\label{figure:tasks-images}
\end{figure}

  All simulation tasks below have a maximum episode length of $T=50$. The episode terminates early if the goal is achieved at a certain step. The sparse binary reward function is $r(s,a,g)=\mathbb{I}[\text{success}]$, which indicates the success of the transitioned state $s^\prime =f(s,a)$\footnote{For such simulation environments, the transition $s^\prime \sim p(\cdot\mid s,a)$ is deterministic so we equivalently write $s^\prime=f(s,a)$ for some deterministic function $f$.}. Below we describe in details the setup of each task, in particular the success criterion.
  \begin{itemize}
  \item \textbf{Point mass \citep{ding2019goal}.} The objective is to navigate a point mass through a 2-D room with obstacles to the target location. $|\mathcal{S}| = 4, |\mathcal{G}| = 2$ and $|\mathcal{A}| = 2$. The goals $g\in\mathbb{R}^2$ are specified as a 2-D point on the plane. Included in the state $s$ are the 2-D coordinates of the point mass, denoted as $s_{xy}\in\mathbb{R}^2$. The success is defined as $d(z(s_{xy}),z(g)) \leq d_0$ where $d(\cdot,\cdot)$ is the Euclidean distance, $z(g)$ is a element-wise normalization function $z(x) \coloneqq (x - x_{\text{min}}) / (x_{\text{max}} - x_{\text{min}})$ where $x_{\text{max}},x_{\text{min}}$ are the boundaries of the wall. The normalized threshold is $d_0 = 0.02 \cdot \sqrt{2}$.
  \item \textbf{Reacher \citep{brockman2016}.} The objective is to move via joint motors the finger tip of a 2-D Reacher robot to reach a target goal location. $|\mathcal{S}| = 11, |\mathcal{G}| = 2$ and $|\mathcal{A}| = 2$. As with the above point mass environment, the goals $g\in\mathbb{R}^2$ are locations of a point at the 2-D plane. Included in the state $s$ are 2-D coordinates of the finger tip location of the Reacher robot $s_{xy}$. The success criterion is defined identically as the point mass environment.
  \item \textbf{Fetch robot \citep{brockman2016,andrychowicz2017hindsight}.} The objective is to move via position controls the end effector of a fetch robot, to reach a target location in the 3-D space. $|\mathcal{S}| = 10, |\mathcal{G}| = 3$ and $|\mathcal{A}| = 3$. This task belongs to the standard environment in OpenAI gym \citep{brockman2016} and we leave the details to the code base and \citep{andrychowicz2017hindsight}.
  \item \textbf{Sawyer robot \citep{nair2018visual,nasiriany2019planning}.} The objective is to move via motor controls of the end effector of a sawyer robot, to reach a target location in the 3-D space. $|\mathcal{S}| = |\mathcal{G}| = |\mathcal{A}| = 3$. This task belongs to the multiworld code base.
  \end{itemize}

 \paragraph{Details on image inputs.} For the customized point mass and Reacher environments, the image inputs are taken by cameras which look vertically down at the systems For the Fetch robot and Sawyer robot environment, the images are taken by cameras mounted to the robotic systems. See \Cref{figure:tasks-images} for an illustration of the image inputs.

 \subsection{Details on Algorithms and Hyper-parameters}

 \paragraph{Hindsight Expectation Maximization.} For \gls{hEM} on domains with discrete actions, the policy is a categorical distribution $\pi_\theta(a\mid s,g) = \text{Cat}(\phi_\theta(s,g))$ with parameterized logits $\phi_\theta(s,g)$; on domains with continuous actions, the policy network is a state-goal conditional Gaussian distribution $\pi_\theta(a\mid s,g) = \mathcal{N}(\mu_\theta(s,g),\sigma^2)$ with a parameterized mean $\mu_\theta(s,g)$ and a global standard deviation $\sigma^2$. The mean is takes the concatenated vector $[x,g]$ as inputs, has $5$ hidden layers each with $5$ hidden units interleaved with $\text{relu}(x)$ non-linear activation functions, and outputs a vector $\mu_\theta(s,g) \in \mathbb{R}^{|\mathcal{A}|}$.

 \gls{hEM} alternates between data collection using the policy and policy optimization with the \gls{EM}-algorithms. During data collection, the output action is perturbed by a Gaussian noise $a^\prime = \mathcal{N}(0,\sigma_a^2) + a, a\sim \pi_\theta(\cdot\mid s,g)$ where the scale is $\sigma_a = 0.5$. Note that injecting noise to actions is a common practice in off-policy \gls{RL} algorithms to ensure sufficient exploration \citep{mnih2013,lillicrap2015continuous}; for tasks with a  discrete action space, the agent samples actions $a\sim\pi_\theta(\cdot\mid s,g)$ with probability $1-\epsilon$ and uniformly random with probability $\epsilon\in[0.2,0.5]$. The baseline \gls{hEM} collects data with $N=30$ parallel MPI actors, each with $k=20$ trajectories. When sampling the hindsight goal given trajectories, we adopt the \emph{future} strategy specified in \gls{HER} \citep{andrychowicz2017hindsight}: in particular, at state $s$, future achieved goals are uniformly sampled at trainig time as $q_h(\tau,g)$. All parameters are optimized with Adam optimizer \citep{kingma2014adam} with learning rate $\alpha=10^{-3}$ and batch size $B=64$. By default, we run $M=30$ parallel MPI workers for data collection and training, at each iteration \gls{hEM} collects $N=20$ trajectories from the environment. For image-based reacher and Fetch robot, \gls{hEM} collects $N=80$ trajectories.


 \paragraph{Hindsight Experience Replay.} By design in \citep{andrychowicz2017hindsight}, \gls{HER} is combined with off-policy learning algorithms such as DQN or DDPG \citep{mnih2013,lillicrap2015continuous}. We describe the details of DDPG. The algorithm maintains a Q-function $Q_\theta(s,a,g)$ parameterized similarly as a universal value function \citep{schaul2015prioritized}: the network takes as inputs the concatenated vector $[x,a,g]$, has $5$ hidden layers with $h=256$ hidden units per layer interleaved with $\text{relu}(x)$ non-linear activation functions, and outputs a single scalar. The policy network $\pi_\theta(s,g)$ takes the concatenated vector $[x,g]$ as inputs, has the same intermediate architecture as the Q-function network and outputs the action vector $\pi_\theta(s,g) \in \mathbb{R}^{|\mathcal{A}|}$. We take the implementation from OpenAI baseline \citep{baselines}, all missing hyper-parameters are the default hyper-parameters in the code base. Across all tasks, \gls{HER} is run with $M=20$ parallel MPI workers as specified in \citep{baselines}.

 \paragraph{Image-based architecture.}
 When state or goal are image-based, the Q-function network/policy network applies a convolutional network to extract features. For example, let $s,g \in \mathbb{R}^{w\times w \times 3}$ where $w\in \{48,84\}$ be raw images, and let $f_\theta(s),f_\theta(g)$ be the features output by the convolutional network. These features are concatenated before passing through the fully-connected networks described above. The convolutional network has the following architecture: $[32,8,4] \rightarrow \text{relu} \rightarrow [64,4,2] \rightarrow \text{relu} \rightarrow [64,3,2] \rightarrow \text{relu}$, where $[n_f,r_f,s_f]$ refers to: $n_f$ number of feature maps, $r_f$ feature patch dimension and $s_f$ the stride.

\subsection{Ablation study}

\begin{figure}[h]
\centering
\subcaptionbox{\textbf{Flip bit}}[.3\linewidth]{\includegraphics[width=1.5in]{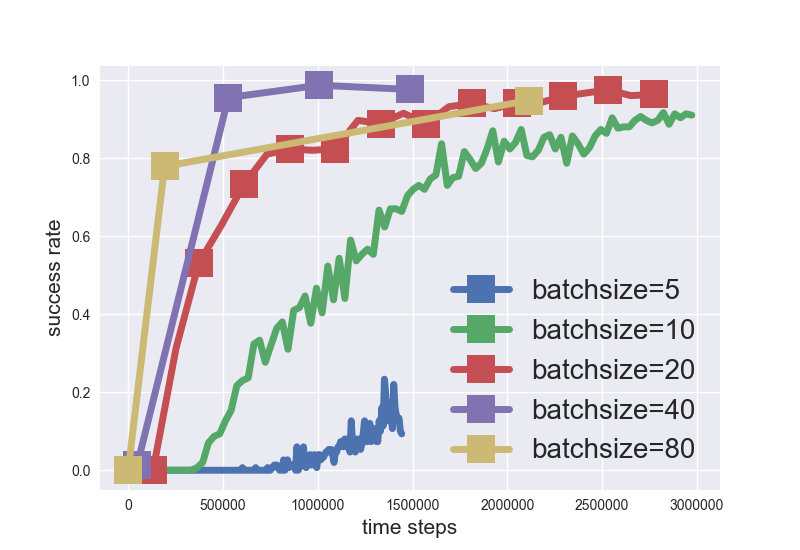}}
\subcaptionbox{\textbf{Fetch robot}}[.3\linewidth]{\includegraphics[width=1.5in]{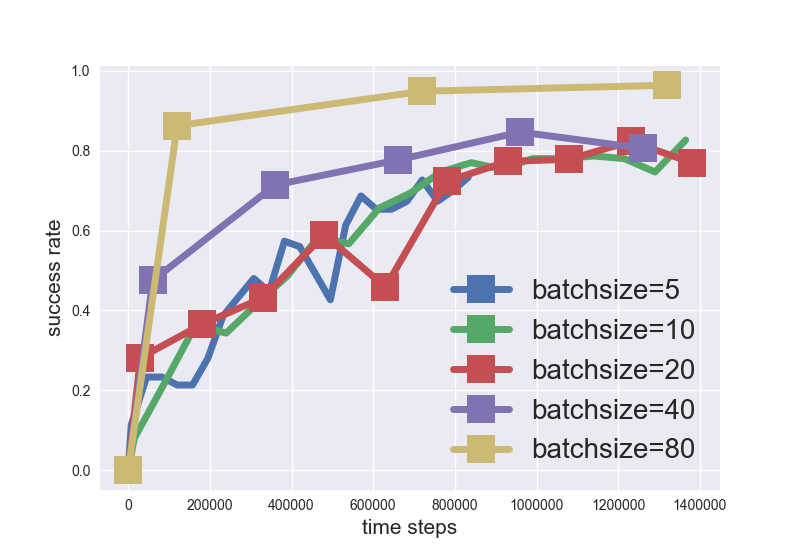}}
\subcaptionbox{\textbf{Sawyer robot}}[.3\linewidth]{\includegraphics[width=1.5in]{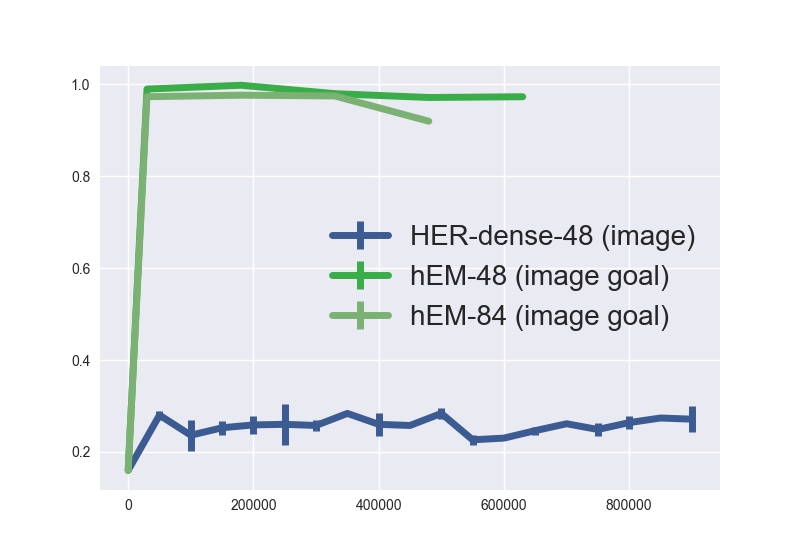}}
\caption{\small{Ablation study. Plot (a) and (b): The effect of the data collection size $N$. Plot (c): The effect of image-based inputs for both states and goals. `\gls{hEM}-48' refers to image-based inputs with size $48 \times 48 \times 3$. Note also `\gls{HER}-dense' is equivalent to `\gls{HER}'($-1/0$) - though the reward function does not provide additional information compared to $(0/1)$, in practice this transformation makes the learning much more stable.}}
\label{figure:ablation-appendix}
\end{figure}

\paragraph{Ablation study on the effect of $N$.} \gls{hEM} collects $N$ trajectories at each training iteration. We vary $N\in\{5,10,20,40,80\}$ on two challenging domains: Flip bit $K=50$ and Fetch robot (image-based) and evaluate the corresponding performance. See \Cref{figure:ablation-appendix}. We see that in general, large $N$ tends to lead to better performance. For example, when $N=5$, \gls{hEM} learns slowly on Flip bit; when $N=80$, \gls{hEM} generally achieves faster convergence and better asymptotic performance across both tasks. We speculate that this is partly because with large $N$ the algorithm can have a larger coverage over goals (larger support over goals in the language of \Cref{thm:monotonic}). With small $N$, the policy might converge prematurely and hence learn slowly. Similar observations have been made for \gls{HER}, where they find that the algorithm performs better with a large number of MPI workers (effectively large $N$).

\paragraph{Ablation on image-based goals.} To further assess the robustness of \gls{hEM} against image-based inputs, we consider Sawyer robot where both states and goals are image-based. This differs from experiments shown in \Cref{figure:robotics-image} where only states are image-based. In \Cref{figure:ablation-appendix}(c), we see that the performance of \gls{hEM} does not degrade even when goals are image-based and is roughly agnostic to the size of the image. Contrast this with \gls{HER}, which does not make significant progress even when only states are image-based.

\subsection{Comparison between \gls{hEM} and \gls{HPG}}


 We do not list \gls{HPG} as a major baseline for comparison in the main paper, primarily due to a few reasons: by design, the \gls{HPG} agent tackles discrete action space (see the author code base \url{https://github.com/paulorauber/hpg}), while many goal-conditioned baselines of interest \citep{andrychowicz2017hindsight,nair2018visual,nasiriany2019planning} are continuous action space. Also, in \citep{rauber2017hindsight} the author did not report comparison to traditional baselines such as \gls{HER} and only report cumulative rewards instead of success rate as evaluation criterion. Here, we compare \gls{hEM} with \gls{HPG} on a few discrete benchmarks provided in \citep{rauber2017hindsight} to assess their performance.

 \paragraph{Details on \gls{HPG}.} The \gls{HPG} is based on the author code base. \citep{rauber2017hindsight} proposes several \gls{HPG} variants with different policy gradient variance reduction techniques \citep{sutton1999} and we take the \gls{HPG} variant with the highest performance as reported in the paper. Throughout the experiment we set the learning rate to be $10^{-3}$ and other hyper-parameters take default values.

 \paragraph{Benchmarks.} We compare \gls{hEM} and \gls{HPG} on Flip bit $K=25,50$ and the four room environment. The details of the Flip bit environment could be found in the main paper. The four room environment is used as a benchmark task in \citep{rauber2017hindsight}, where the agent navigates a grid world with four rooms to reach a target location within episodic time limit. The agent has access to four actions, which moves the agent in four directions. The trial is successful only if the agent reaches the goal in time.

\begin{figure}[h]
\centering
\subcaptionbox{\textbf{Flip bit $K=25$}}[.3\linewidth]{\includegraphics[width=1.5in]{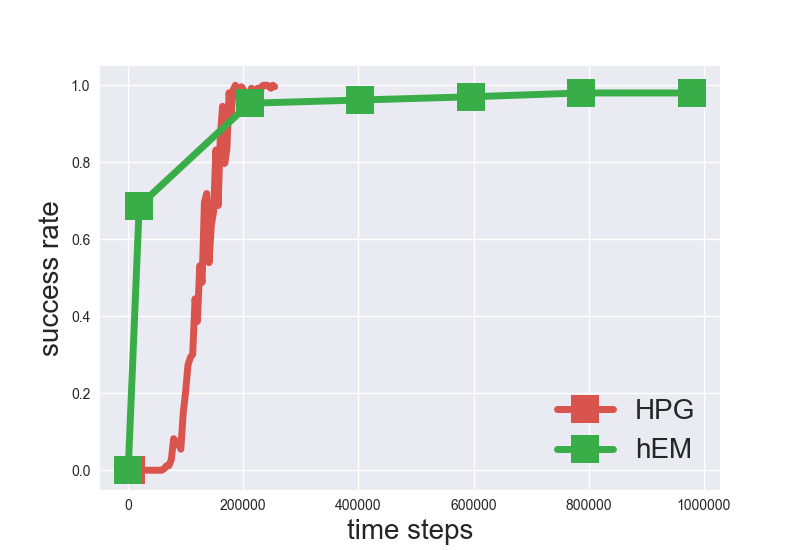}}
\subcaptionbox{\textbf{Flip bit $K=50$}}[.3\linewidth]{\includegraphics[width=1.5in]{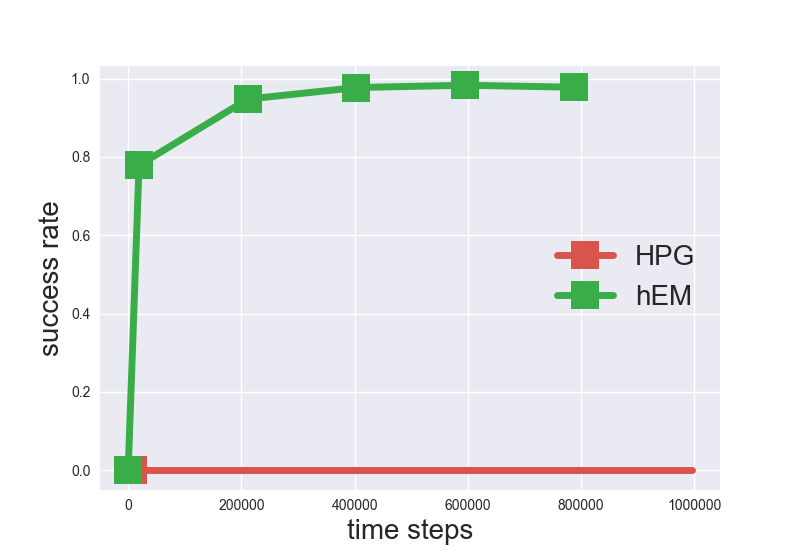}}
\subcaptionbox{\textbf{Four room}}[.3\linewidth]{\includegraphics[width=1.5in]{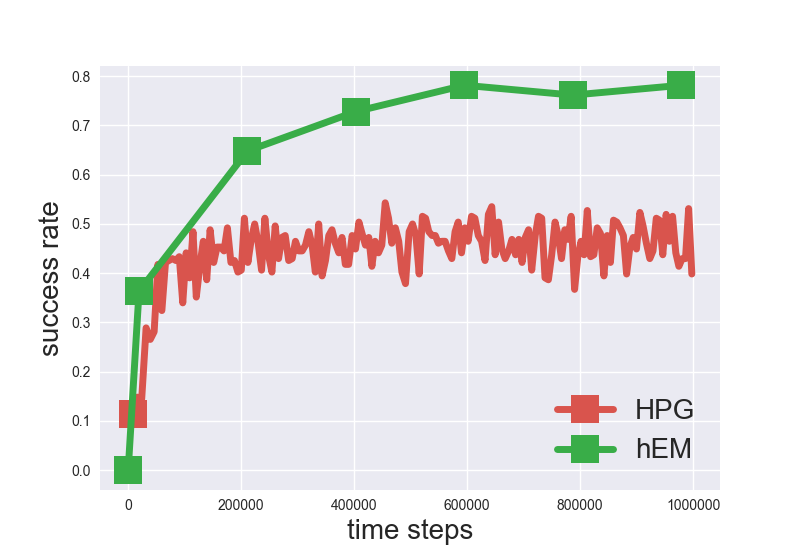}}
\caption{\small{Comparison between \gls{hEM} and \gls{HPG}. \gls{HPG} performs well on Flip bit \gls{MDP} with $K=25$, but when $K=50$ its performance drops drastically. \gls{HPG} also underperforms \gls{hEM} on the four room environment where it makes fast progress initially but saturates to a low sub-optimal level.}}
\label{figure:hpg}
\end{figure}

 \paragraph{Results.} We show results in \Cref{figure:hpg}. For the Flip bit $K=25$, \gls{HPG} and \gls{hEM} behave similarly: both algorithms reach the near-optimal performance quickly and has similar convergence speed; when the state space increases to $K=50$, \gls{HPG} does not make any progress while the performance of \gls{hEM} steadily improves. Finally, for the four room environment, we see that though the performance of \gls{HPG} initially increases quickly as \gls{hEM}, its success rate quickly saturates to a level significantly below the asymtotpic performance of \gls{hEM}. These observations show that \gls{hEM} performs much more robustly and significantly better than \gls{HPG}, especially in challenging environments.

\subsection{Additional Results on Baseline Fetch Tasks}

\paragraph{Details on Fetch tasks.} Fetch tasks
are introduced in \citep{andrychowicz2017hindsight}: \textbf{Reach}, \textbf{Slide}, \textbf{Push} and \textbf{Pick and Place}. We have already evaluated on the Reach task in our prior experiments. Here, we focus on the other three experiments.

\paragraph{Issues with exploration.} As we have alluded to in the main paper, exploration is a critical issue in goal-conditioned \gls{RL}, as exemplified through the Fetch tasks. Compared to the reaching tasks we considered, Fetch tasks in general have more complicated goal space - purely random exploration might not cover the entire goal space fully. This observation has been corroborated in recent work \citep{zhang2020automatic,pitis2020maximum}.

\paragraph{Details on algorithms.} To solve these tasks, we build on the built-in exploration mechanism of \gls{HER} implementations \citep{baselines}, which have been shown to work in certain setups. Though \gls{hEM} does not utilize any value function critics, it shares the policy network with \gls{HER}. We implement the aggregate loss function as
\begin{align*}
    L(\theta) = L_\text{her} + \eta \cdot L_\text{hem} =  &\mathbb{E}_{(s,g)\sim\mathcal{D}} \left[Q_\phi \left(s,\pi_\theta(s),g\right)\right]+ \eta \cdot \mathbb{E}_{q(\tau,g)}\left[\sum_{t=0}^{T-1} \log \pi_\theta(a_t \mid s_t,g) \right].
\end{align*}
 for the policy network. Here $\eta\in\{0.1, 0.2,0.5\}$ is a hyper-parameter we selected manually to determine the trade-offs between two loss functions. Intuitively, when the policy cannot benefit from the learning signals of \gls{HER} via $L_{\text{her}}$, it is still able to learn via the supervised learning update through $L_{\text{hem}}$. In practice, we find $\eta=0.1$ works uniformly well.

\begin{figure}[h]
\centering
\subcaptionbox{\textbf{ Slide}}[.3\linewidth]{\includegraphics[width=1.5in]{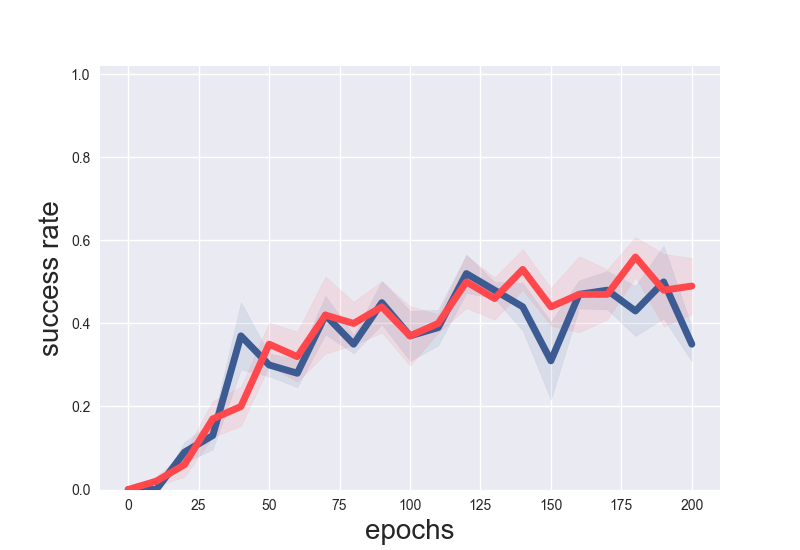}}
\subcaptionbox{\textbf{ Push}}[.3\linewidth]{\includegraphics[width=1.5in]{plots/fetchpush.png}}
\subcaptionbox{\textbf{Pick and Place}}[.3\linewidth]{\includegraphics[width=1.5in]{plots/fetchpickandplace.png}}
\caption{\small{Fetch tasks. Training curves of \gls{hEM} and \gls{HER} on two standard Fetch tasks. \gls{hEM} provides marginal speed up compared to \gls{HER}. All curves are calculated based on $5$ random seeds.}}
\label{figure:fetch-appendix}
\end{figure}

 \paragraph{Results.} See \Cref{figure:fetch-appendix} for the full results. Note that \gls{hEM} provides marginal speed up on the learning on Push and Pick \& Place. On Slide, both algorithms get stuck at local optimal (similar observations were made in \citep{andrychowicz2017hindsight}). We speculate that improvements in the exploration literature would bring further consistent performance gains.

\end{document}